\documentclass{osa-article}

\journal{boe}

\usepackage[utf8x]{inputenc}

\begin{document}

\title{Dynamic Deep Networks for Retinal Vessel Segmentation}

\author{Aashis Khanal\authormark{1,2} and Rolando Estrada\authormark{1,3}}

\address{\authormark{1}Department of Computer Science, Georgia State University, Atlanta, GA 30303, USA\\
\authormark{2}akhanal1@student.gsu.edu\\
\authormark{3}restrada1@gsu.edu
}




\begin{abstract*}
Segmenting the retinal vasculature entails a trade-off between how much of the overall vascular structure we identify vs. how precisely we segment individual vessels. In particular, state-of-the-art methods tend to under-segment faint vessels, as well as pixels that lie on the edges of thicker vessels. Thus, they underestimate the width of individual vessels, as well as the ratio of large to small vessels. More generally, many crucial bio-markers---including the artery-vein (AV) ratio, branching angles, number of bifurcation, fractal dimension, tortuosity, vascular length-to-diameter ratio and wall-to-lumen length---require precise measurements of individual vessels. To address this limitation, we propose a novel, stochastic training scheme for deep neural networks that better classifies the faint, ambiguous regions of the image. Our approach relies on two key innovations. First, we train our deep networks with dynamic weights that fluctuate during each training iteration. This stochastic approach forces the network to learn a mapping that robustly balances precision and recall. Second, we decouple the segmentation process into two steps. In the first half of our pipeline, we estimate the likelihood of every pixel and then use these likelihoods to segment pixels that are clearly vessel or background. In the latter part of our pipeline, we use a second network to classify the ambiguous regions in the image. Our proposed method obtained state-of-the-art results on five retinal datasets---DRIVE, STARE, CHASE-DB, AV-WIDE, and VEVIO---by learning a robust balance between false positive and false negative rates. In addition, we are the first to report segmentation results on the AV-WIDE dataset, and we have made the ground-truth annotations for this dataset publicly available.
\end{abstract*}







\section{Introduction}
Retinal vessels provide the only non-invasive view of the cardiovascular system. Thus, they are a key diagnostic feature for a number of diseases, including diabetic retinopathy \cite{doi:10.1117/12.708469}, coronary heart disease \cite{10.1093/eurheartj/ehm221}, and atherosclerosis \cite{10.1167/iovs.03-1390}. However, the current standard of care requires manual inspection by an ophthalmologist, which makes it more challenging for people in developing nations and low-income communities to receive care. For example, only 30\% of African Americans in southern Los Angeles reported being screened for diabetic retinopathy, despite being the most at-risk ethnicity \cite{Lue31}. Therefore, it is vital to develop automatic retinal analysis methods to improve screening rates and public health outcomes.

The first step in a retinal analysis pipeline is to segment the regions in the image that correspond to the vasculature. Formally, vessel segmentation is a binary classification problem, but it presents unique challenges. First, existing datasets are small---typically on the order of 20-40 images---because an ophthalmologist has to manually trace every vessel in each fundus image. In addition, we need to classify hundreds of thousands or even millions of pixels per image, and the labels of nearby pixels are correlated.

Deep neural networks are the state of the art for a wide range of classification problems, but, due to the aforementioned challenges, training a deep network to segment retinal images is not straightforward. The two main strategies for applying deep learning to this domain are: (1) dividing each image into small patches to maximize the number of training samples \cite{NIPS2012_4741} or (2) combining traditional convolutional layers with upsampling to learn both local and global features \cite{Ronneberger2015UNetCN}. In particular, the U-net architecture \cite{Ronneberger2015UNetCN} and its extensions (e.g., Recurrent U-net \cite{Alom2018RecurrentRC}) have achieved state-of-the-art results by applying the latter strategy. 

However, U-net and other deep learning-based methods favor precision over recall, i.e., they tend to classify ambiguous pixels as background. This strategy is statistically sound since retinal images have about a nine to one ratio of background to vessel pixels, but it under-segments faint vessels, as well as pixels that lie on the edges of thicker vessels. Thus, current methods underestimate the width of individual vessels, as well as the ratio of large to small vessels. This, in turn, compromises later diagnoses since many crucial bio-markers---including the artery-vein (AV) ratio, branching angles, number of bifurcation, fractal dimension, tortuosity, vascular length-to-diameter ratio and wall-to-lumen length---require precise measurements of individual vessels.

\begin{figure}[t]
\includegraphics[width=1\textwidth]{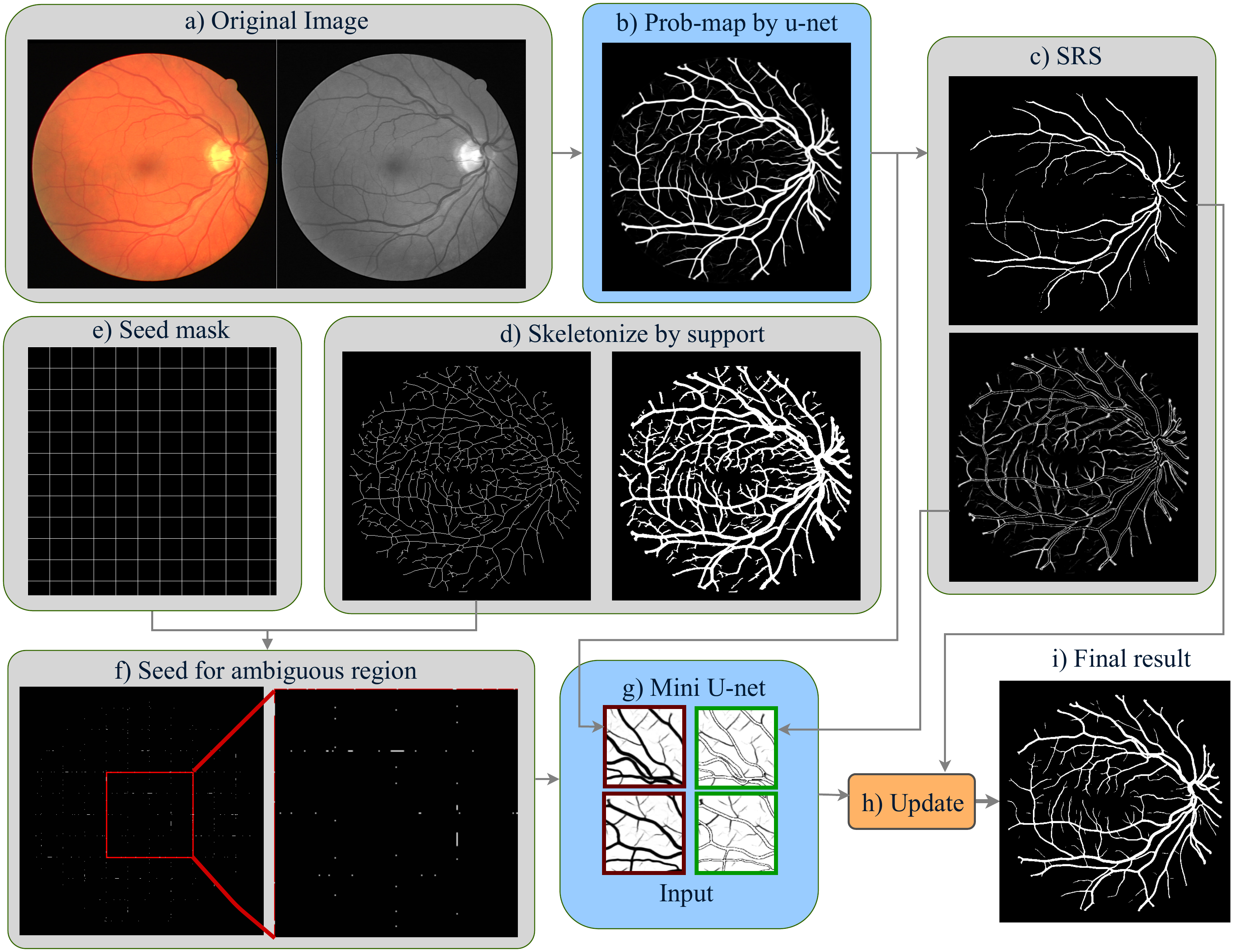}
\caption{{\bf Dynamic Vessel Segmentation Pipeline:} We separate vessel segmentation into simpler tasks. The steps in gray boxes are fixed, while the ones in blue require training. In \textbf{(a)}, we first extract the green channel and preprocess it with contrast adaptive histogram equalization. Then, in \textbf{(b)} we train a U-net architecture using a dynamic loss function. This network outputs a likelihood map across the entire image. In \textbf{(c)}, we use two thresholds, \textit{support} and \textit{resistance}, (SRS) to separate \textit{easy} (top image) from \textit{ambiguous} (bottom image) pixels. We assign the final labels to easy pixels at this stage. To classify ambiguous pixels, in \textbf{(d)} we use the \textit{support} threshold to generate a high-recall estimate of the vascular structure (right image) and then skeletonize this mask (left image). We then intersect this skeleton with a lattice mask (shown in \textbf{(e)}) to find a set of seed points (shown in \textbf{(f)}. The left image in \textbf{(f)} shows the seed locations for the entire image, while the right image is a close-up of the square shown in red. We then define a small patch around each seed point and train a small network (\textit{mini-U-net}) on these patches \textbf{(g)}. Each patch has two channels. The channel highlighted with the purple border is extracted from the original likelihood map, while the channel with the green border includes only the ambiguous pixels in that region. In \textbf{(h)}, we merge the mini-U-net predictions for ambiguous pixels with the labels for the easy pixels to obtain our complete segmentation result (shown in \textbf{(i)}). Figure best viewed on-screen.}
\label{flow_ature}
\end{figure}

In this paper, we propose a novel, two-stage segmentation method that combines both upsampling and patch-based processing. Our approach also leverages loss functions with stochastic weights to better detect faint and ambiguous vessel pixels. As our experiments show, our proposed pipeline achieves state-of-the-art results on five vessel segmentation datasets, including conventional fundus images \cite{drive_dataset,845178,6224174}, wide field-of-view (FOV) images \cite{wide-dataset:6987362}, and low-resolution video stills and mosaics \cite{vevio-dataset:2012}. 


As Fig.~\ref{flow_ature} shows, we separate the overall vessel segmentation problem into two stages: (1) \textit{likelihood estimation} and (2) \textit{targeted prediction}. In the first half of our pipeline, we train a U-net architecture to estimate the vessel likelihood of each pixel using a stochastic cross entropy loss. That is, we randomly assign a weight to each class on each training iteration, which force the network to alternate between higher false positive vs. higher false negative rates during training, preventing it from getting stuck in a single minimum. In the second half of our pipeline, we first separate potential vessel pixels into two categories---\textit{easy} and \textit{ambiguous}---and then train a patch-based network to classify the latter. (Easy pixels can be classified with a simple threshold.) This second network, which we refer to as \textit{mini-U-net}, is a scaled-down version of the network used in the first stage. To the best of our knowledge, we are the first to apply stochastic weights for vessel segmentation, as well as the first to combine both upsampling and patch-based learning into a single pipeline. 







The rest of this paper is organized as follows. First, we discuss related work on vessel segmentation, particularly deep learning-based approaches. Then, we detail our complete methodology, including our stochastic loss function and targeted prediction steps. We then present and discuss experimental results on five datasets and explore avenues for future work.

\section{Related work}
Vessel segmentation has a long history, although the advent of deep neural networks has yielded significant improvements in recent years. Earlier techniques relied primarily on handcrafted features, including matched filters \cite{34715}, quadrature filters \cite{LATHEN2010762}, and Gabor filters \cite{gabor:5929708, gabor1:6729766}. The latter approach, in particular, uses a predefined kernel bank to incorporate all vessel widths and orientations. However, handcrafted features are limited by our ability to analytically model the segmentation process. Other classic techniques, as surveyed further in ~\cite{Kaur2014VariousIS}, include piece-wise thresholding \cite{845178}, region growing \cite{10.1007/10704282_10}, and concavity measurements \cite{5433078}. In addition to handcrafted features, some prior approaches used graph-theoretic techniques to trace the vascular structure, including shortest-path tracking ~\cite{vevio-dataset:2012} and the fast marching algorithm \cite{Benmansour2011}.


Since the breakthrough by convolutional neural networks (CNNs) on the ImageNet dataset in 2012 ~\cite{imagenet:NIPS2012_4824}, deep learning has achieved tremendous success across a wide array of machine learning tasks ~\cite{imagenet:NIPS2012_4824, r2unet:journals/corr/abs-1802-06955, con_gan_unet:journals/corr/abs-1805-04224, DBLP-maskrcnn:journals/corr/HeGDG17}. Currently, deep networks are the state of the art in vessel segmentation. In particular, the most successful deep architecture in this domain---U-net \cite{Ronneberger2015UNetCN}---combines traditional convolutional layers with feature upsampling to estimate both local and global image features. As noted above, earlier deep learning approaches \cite{patch_based1:7950512, pixel_patch_based2:NIPS2012_4741} divided an image into a series of patches and applied a classifier to each patch independently. For instance, Dasgupta et al. divided the full image into small patches to train a CNN ~\cite{patch_based1:7950512}, while Ciresan et al. defined a patch around each pixel and used those patches to train their network ~\cite{pixel_patch_based2:NIPS2012_4741}. One of the drawbacks of patch-based approaches is that we need to train a CNN with a large number of patches, most of which have redundant information. The upsampling in U-net, on the other hand, allows a network to train with fewer, significantly larger patches. 

There are numerous extensions to the original U-net architecture. Alom et al. increased the performance of U-net by introducing residual connection, and a recursive training strategy ~\cite{r2unet:journals/corr/abs-1802-06955}, albeit at the expense of increased training time due to the extra connections and recursion. Zhuang et al. arranged two U-net architectures in serial fashion, enabling the architecture to learn more abstract features ~\cite{laddernet:journals/corr/abs-1810-07810}. This serialization significantly increases the training time and requires heavy computational infrastructure to be able to train well. Jin \textit{et al.} used deformable CNNs to construct better vascular features \cite{2018arXiv181101206J}, but also with added training complexity. Finally, Yun \textit{et al.} combined conditional generative adversarial networks (GANs) with U-net \cite{con_gan_unet:journals/corr/abs-1805-04224}, achieving results comparable to the other U-net extensions. In contrast to these extensions, our approach not only yielded better results, but is simpler and faster to train, as detailed in the following section.


\section{Methodology}  
As noted above, we separate vessel segmentation into two stages. Fig.~\ref{flow_ature} details each step of our pipeline. In the first stage, we use a stochastic loss function to obtain an overall \textit{likelihood map} for every pixel in the image. Then, we identify those pixels that are most likely to be misclassified---generally faint vessels and pixels at the periphery of thick vessels---and apply a second, smaller network to these ambiguous regions.  Below, we first present and motivate the use of stochastic loss functions and then discuss each step in our pipeline in turn.

\subsection{Dynamic loss functions}
A weighted loss function penalizes some types of errors more than others. They are useful for problems with unbalanced datasets or in which we want to prioritize certain outcomes (e.g., minimize false positives for a particular class). Traditionally, these weights have been fixed \textit{a priori} by estimating them using a training set. 

In this work, however, we propose using \textit{dynamic} or stochastic weights. In this case, each weight is defined not as a single value but as a distribution. During training, every time we feed a new mini-batch to the network, we first sample weights for each class and then use those sampled weights to compute the loss. Since the weights will vary for each training iteration, the same error will be penalized more or less heavily at different points during training. Empirically, this forces the network to optimize across \textit{all} ranges of false positive vs. false negative rates, leading to more robust classification. For concreteness, below we discuss the use of dynamic weights for the cross-entropy loss, but one can randomize any weighted loss function. 

The cross-entropy loss is a natural choice for binary classification since it encourages values to converge towards zero or one. This loss is defined as follows:
\begin{equation}
	\label{eq:cross_entrpy}
	H = -\sum_{x}p(x)\log{q(x)}
\end{equation} 
where $p$ is the true distribution and $q$ is the predicted distribution, i.e., the network's outputs. The training process aims to make these two distributions as close as possible. However, when the data is highly skewed, we can trivially obtain a high accuracy by always assigning the most common class to every pixel. In the case of vessel segmentation, very few pixels are part of a vessel, so always assigning a label of \textit{background} will yield an accuracy of around 90\% (but a recall of 0\%). A straightforward way to ameliorate this imbalance is by applying different weights to each class:
\begin{equation}
	\label{eq:weighted_cross_entrpy}
	H = -\sum_{x}w{_i}p(x)\log{q(x)}
\end{equation}
where $w_i$ are the class weights and $i={1,...,C}$ are the class labels ($C = 2$ for binary segmentation). By using weights, we force the network to care about both positive and negative samples equally during training. An easy way to define these weights is to estimate them from the training set, as was done for the original U-net architecture \cite{Ronneberger2015UNetCN}.

A major drawback of the above strategy is that it imposes a fixed ratio of vessel-to-background. As our experiments show, this ratio is a major determinant of whether ambiguous pixels will be mapped to either vessel or background. In other words, when faced with an ambiguous pixel, the network will use this ratio to "guess" which label to assign to it. We can see in Figs.~\ref{fig:pmap_datasets}(b) and \ref{fig:pmap_datasets}(c) that a network trained with fixed weights will preferentially over- or under-segment ambiguous pixels.

Instead, we propose using stochastic weights which are randomly generated at each training iteration: 
\begin{equation}
	\label{eq:dynamic_cross_entrpy}
	H = -\sum_{x}w{_{rand(1, \alpha, s)}}p(x)\log{q(x)}
\end{equation}
where function $rand(1, \alpha, s)$ assigns random class weight within a range of $1$ to $alpha$ and $s$ is the step by which the random class weights are taken. For example, $s$ = 1 would generate random integers between $[1, \alpha]$. Intuitively, dynamic weights prevent the network from getting stuck in a low-value local minimum. 

When we use fixed class weights, the neural network will always penalize all pixels by the same factor. As noted earlier and as Figs.~\ref{fig:pmap_datasets}(b) and \ref{fig:pmap_datasets}(c) show, fixed weights fail to account for fainter vessels and are insensitive to noise. Stochastic costs reduce this bias because the network must vary in how aggressively it segments ambiguous pixels throughout the training process. Even if completely different weights are given in a subsequent iteration, some of the information learned in the previous step is preserved, leading to a more balanced segmentation of ambiguous regions. As Fig.~\ref{fig:pmap_datasets}(d) shows, dynamic weights capture fainter vessels more accurately than fixed weights.

We now discuss the various steps of our proposed pipeline below.

\subsection{Likelihood map}
The first step in our pipeline consists of training a U-net architecture using stochastic weights to derive a likelihood map over the entire image. The goal of this map is not to segment the image directly but to indicate how likely a particular pixel is to be part of a vessel. We do not segment the image at this stage because the up-sampling used by U-net blurs the original vessels, making it difficult to correctly label high-frequency elements (e.g., thin vessels or pixels near the edge of a thick vessel).

\begin{figure}
    \begin{minipage}{0.243\textwidth}
        \centering
        \includegraphics[width=\textwidth]{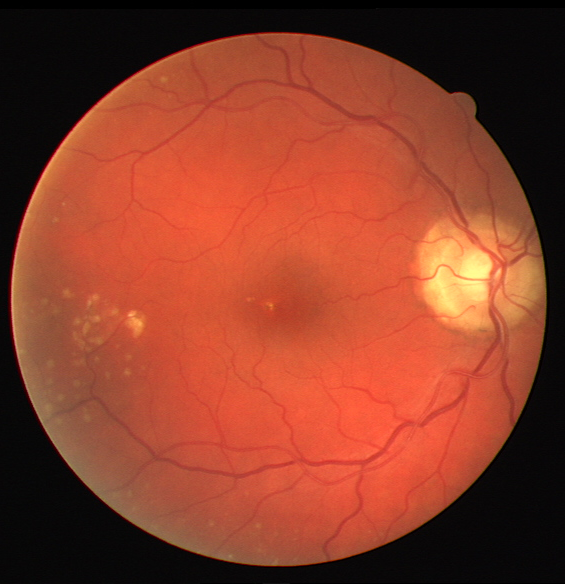}
    \end{minipage}
    \hfill
    \begin{minipage}{0.243\textwidth}
        \centering
        \includegraphics[width=\textwidth]{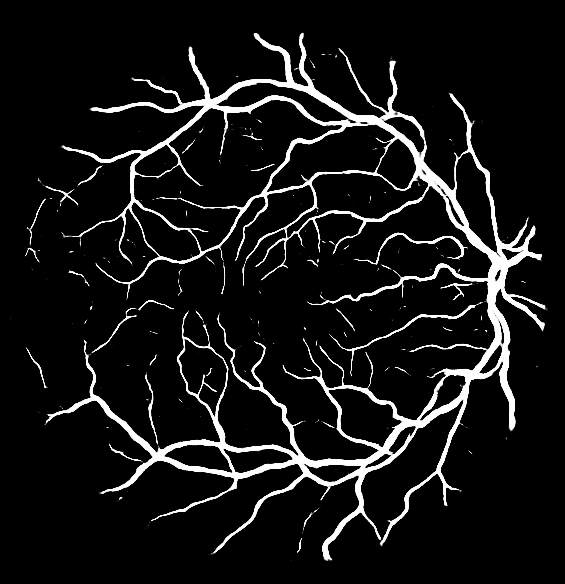}
    \end{minipage}
    \hfill
    \begin{minipage}{0.243\textwidth}
        \centering
        \includegraphics[width=\textwidth]{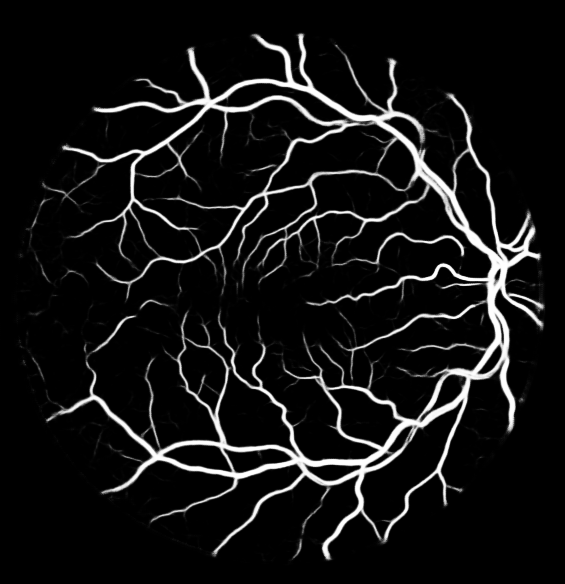}
    \end{minipage}
    \hfill
    \begin{minipage}{0.243\textwidth}
        \centering
        \includegraphics[width=\textwidth]{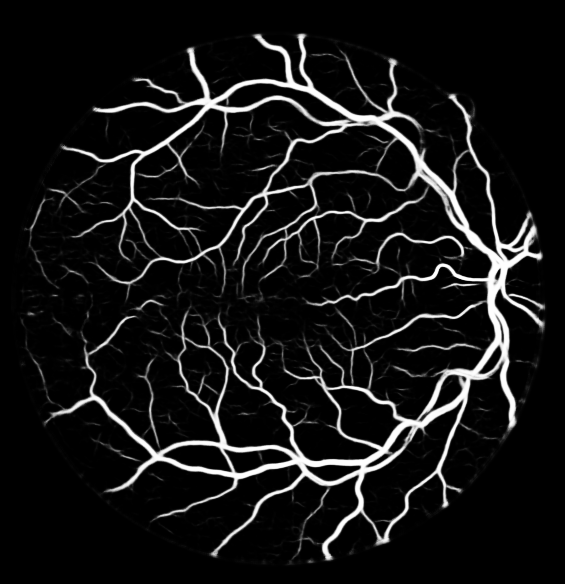}
    \end{minipage}

    \begin{minipage}{0.243\textwidth}
        \centering
        \includegraphics[width=\textwidth]{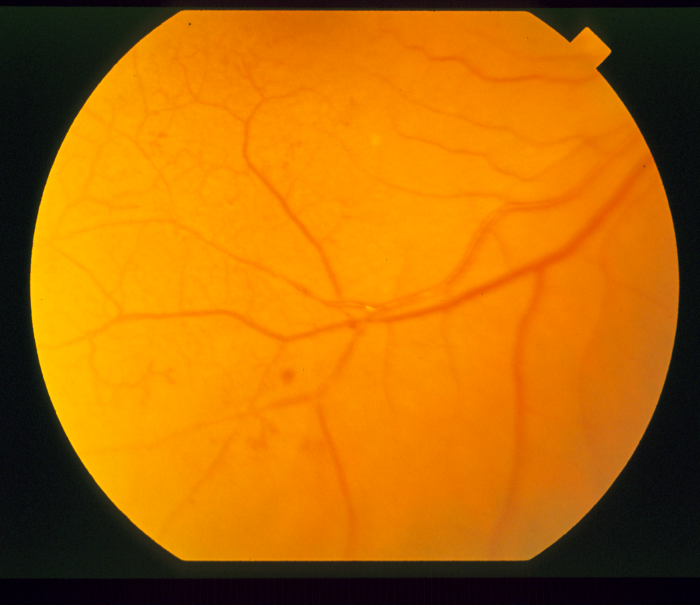}
    \end{minipage}
    \hfill
    \begin{minipage}{0.243\textwidth}
        \centering
        \includegraphics[width=\textwidth]{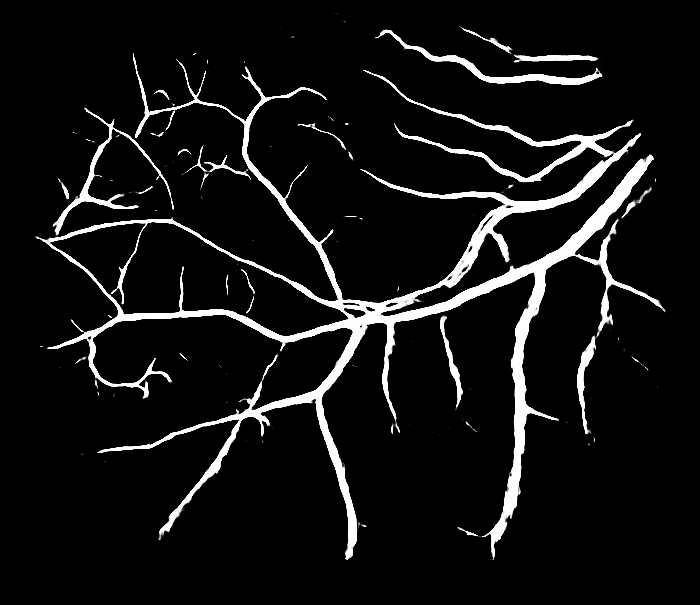}
    \end{minipage}
    \hfill
    \begin{minipage}{0.243\textwidth}
        \centering
        \includegraphics[width=\textwidth]{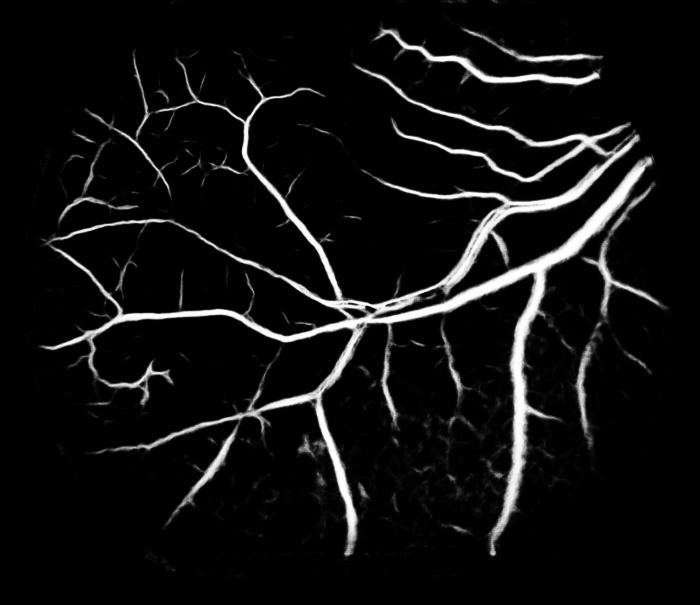}
    \end{minipage}
    \hfill
    \begin{minipage}{0.243\textwidth}
        \centering
        \includegraphics[width=\textwidth]{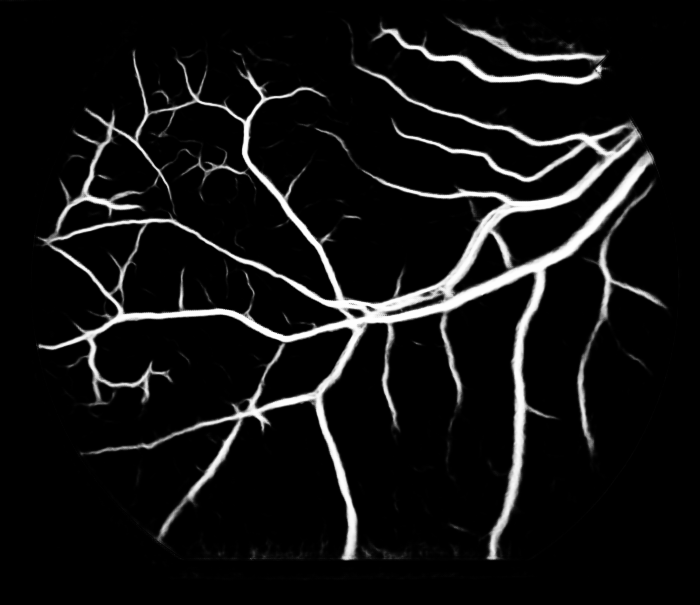}
    \end{minipage}
    
    \begin{minipage}{0.243\textwidth}
        \centering
        \includegraphics[width=\textwidth]{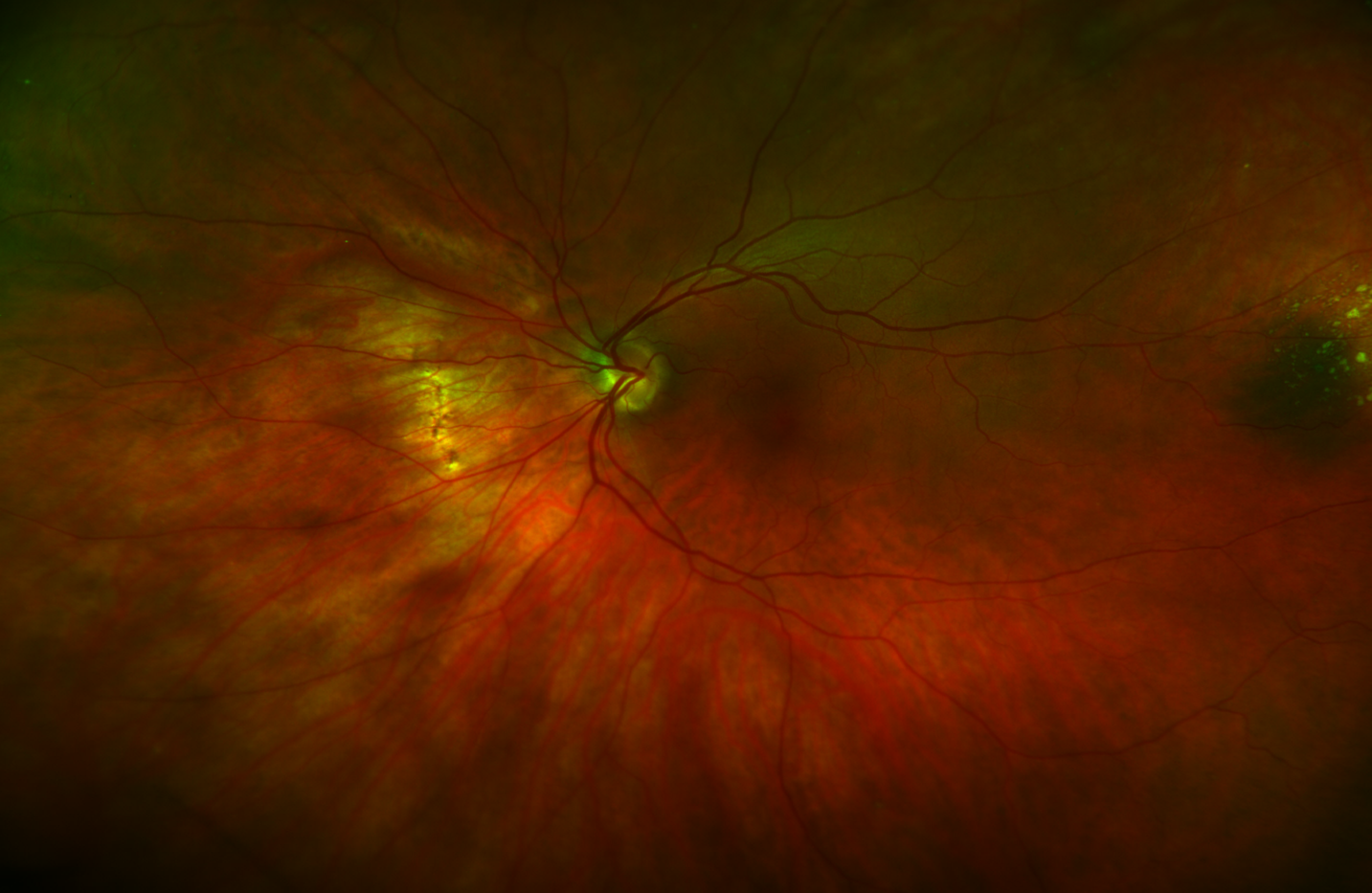}
    \end{minipage}
    \hfill
    \begin{minipage}{0.243\textwidth}
        \centering
        \includegraphics[width=\textwidth]{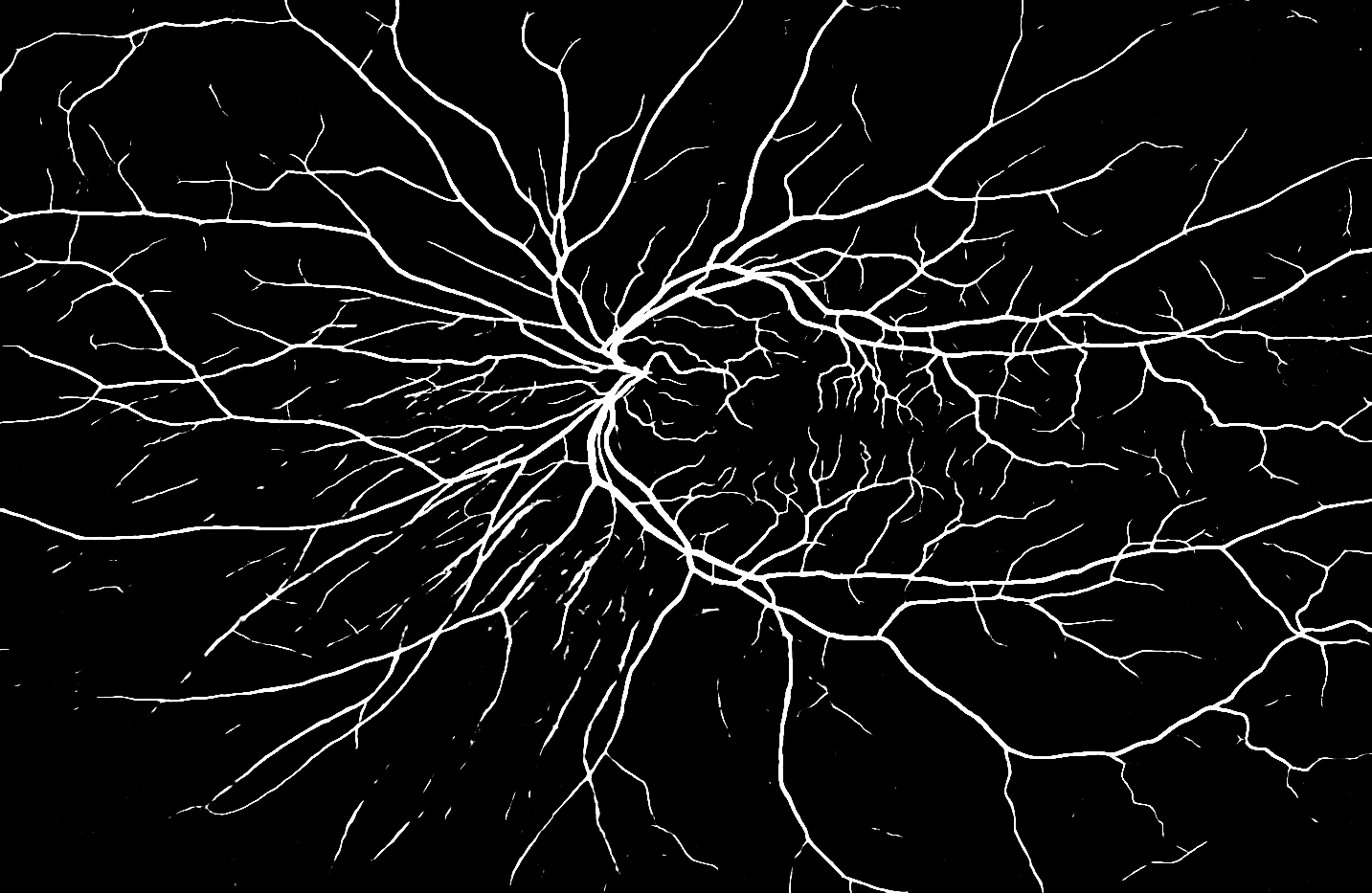}
    \end{minipage}
    \hfill
    \begin{minipage}{0.243\textwidth}
        \centering
        \includegraphics[width=\textwidth]{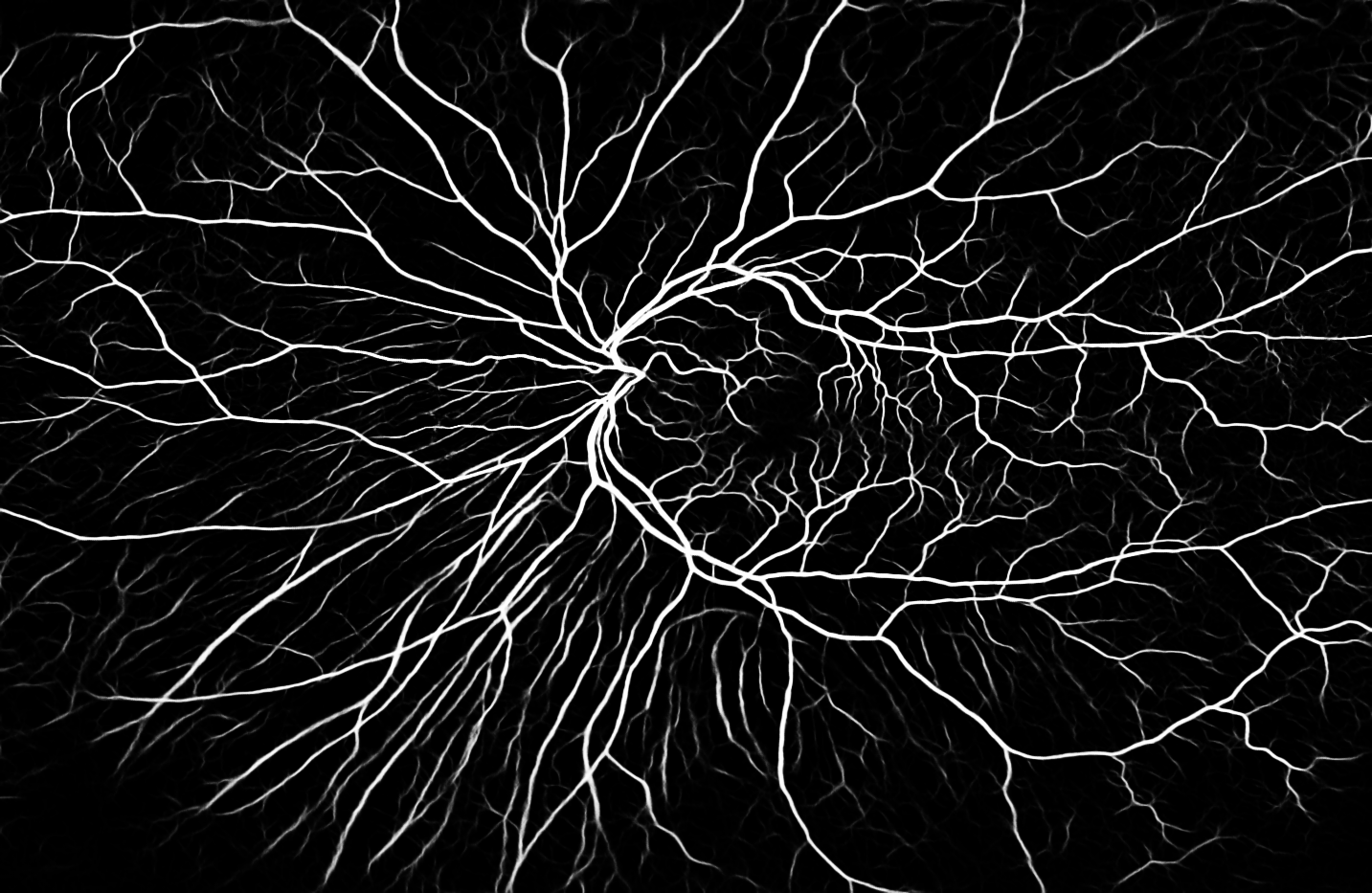}
    \end{minipage}
    \hfill
    \begin{minipage}{0.243\textwidth}
        \centering
        \includegraphics[width=\textwidth]{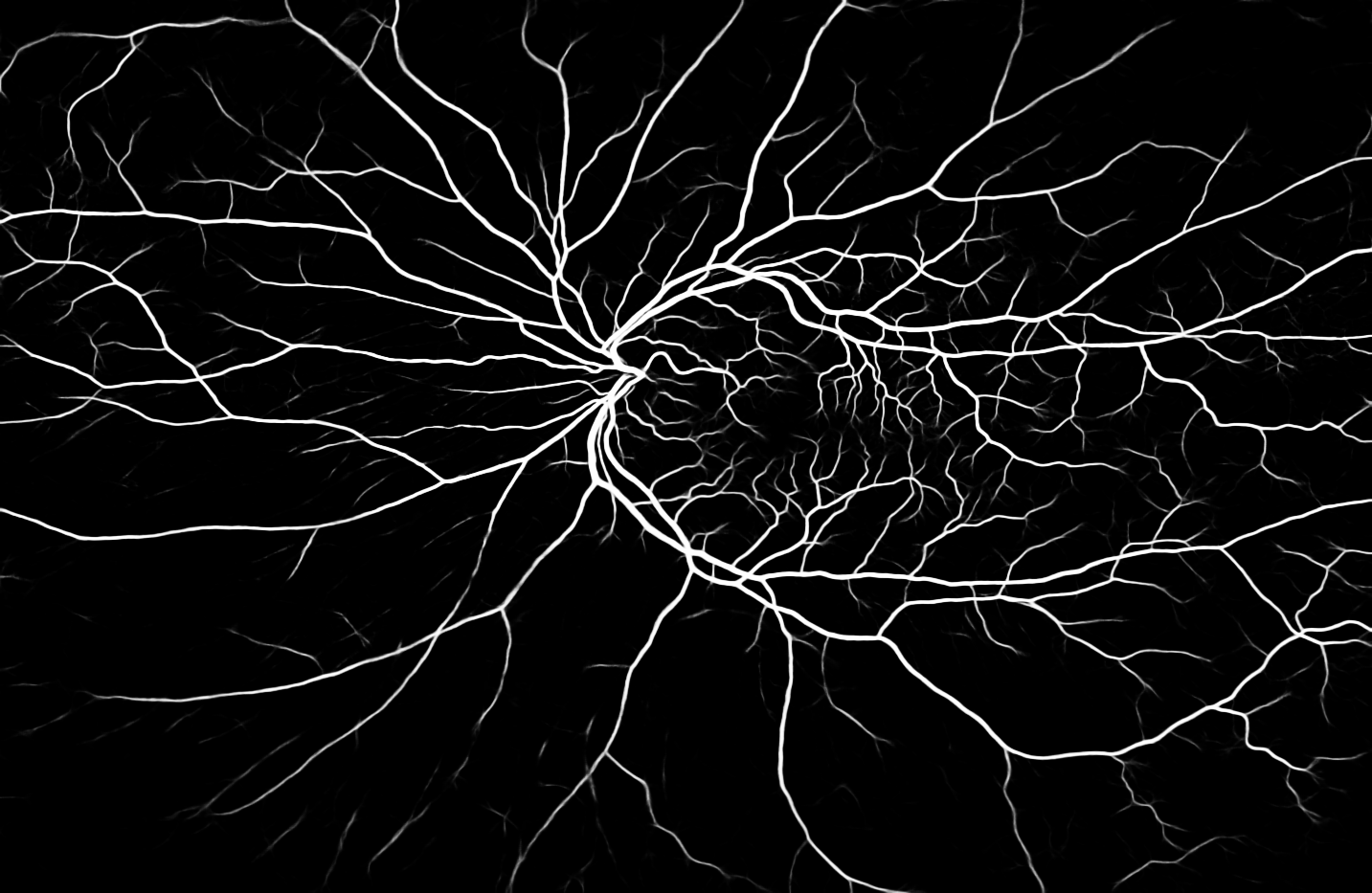}
    \end{minipage}
    
    \begin{minipage}{0.243\textwidth}
        \centering
        \includegraphics[width=\textwidth]{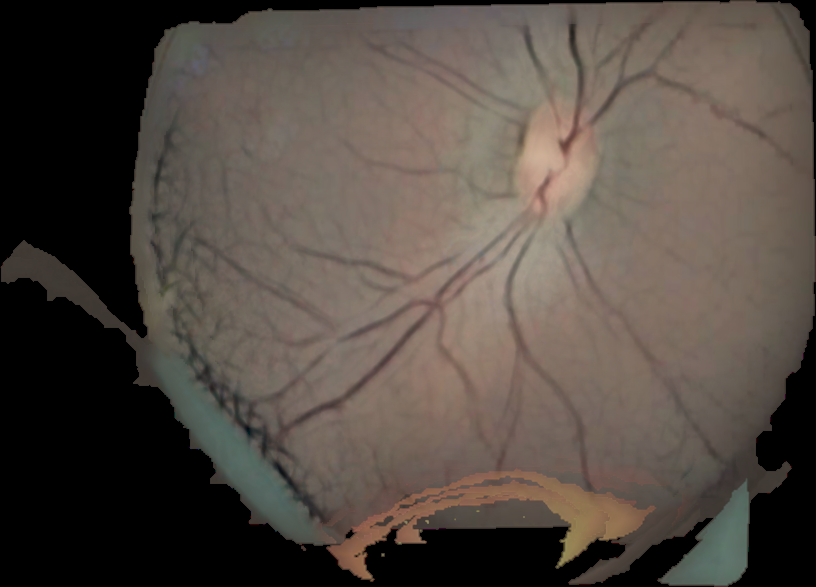}
    \end{minipage}
    \hfill
    \begin{minipage}{0.243\textwidth}
        \centering
        \includegraphics[width=\textwidth]{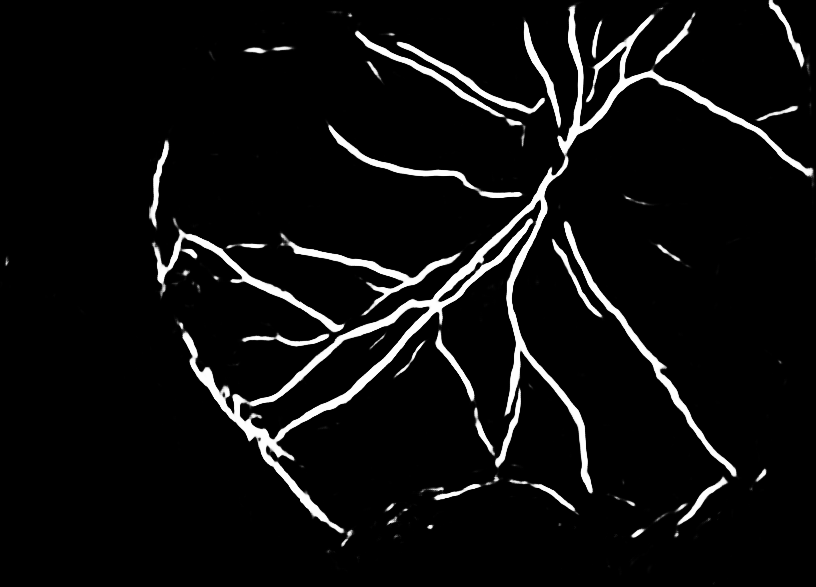}
    \end{minipage}
    \hfill
    \begin{minipage}{0.243\textwidth}
        \centering
        \includegraphics[width=\textwidth]{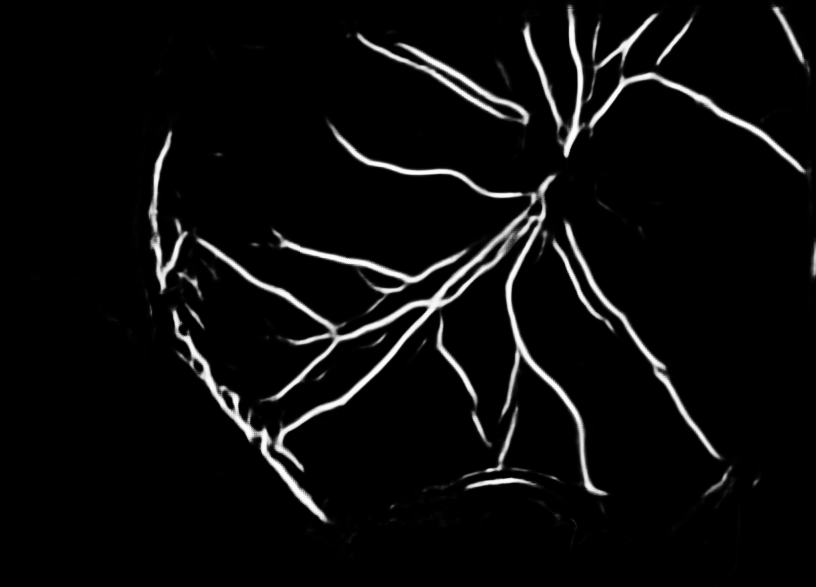}
    \end{minipage}
    \hfill
    \begin{minipage}{0.243\textwidth}
        \centering
        \includegraphics[width=\textwidth]{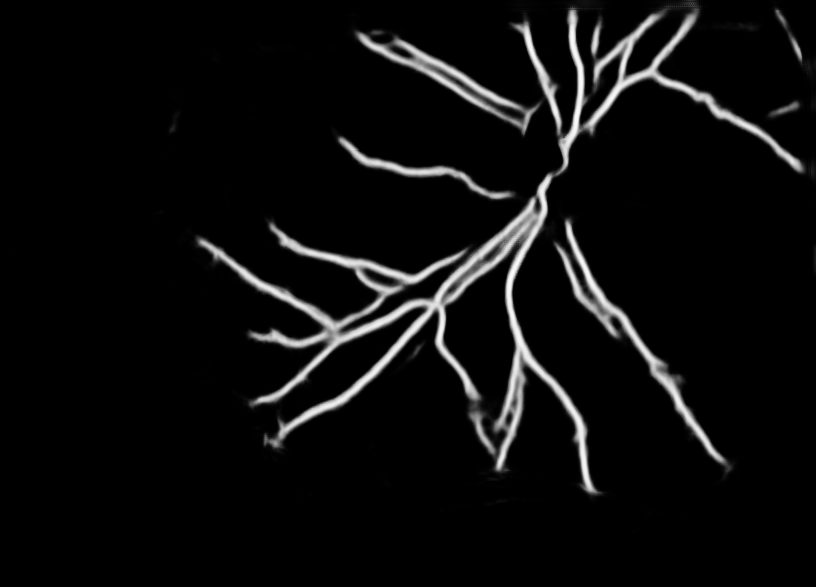}
    \end{minipage}
    
    \begin{minipage}{0.243\textwidth}
        \centering
        \includegraphics[width=\textwidth]{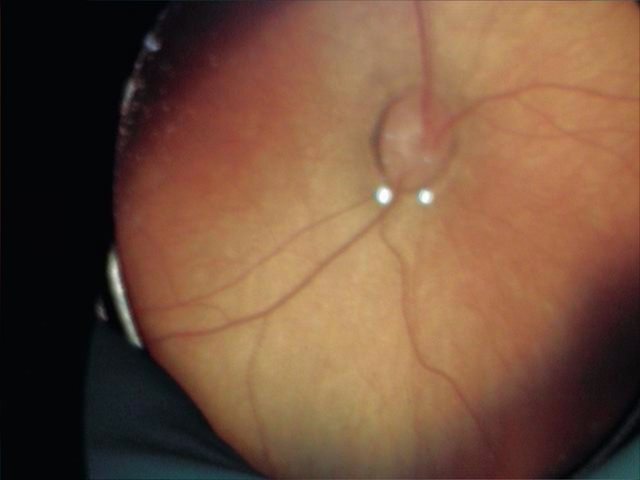}
    \end{minipage}
    \hfill
    \begin{minipage}{0.243\textwidth}
        \centering
        \includegraphics[width=\textwidth]{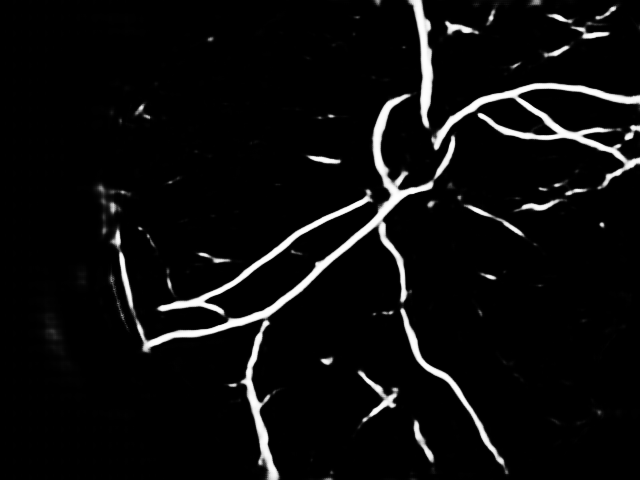}
    \end{minipage}
    \hfill
    \begin{minipage}{0.243\textwidth}
        \centering
        \includegraphics[width=\textwidth]{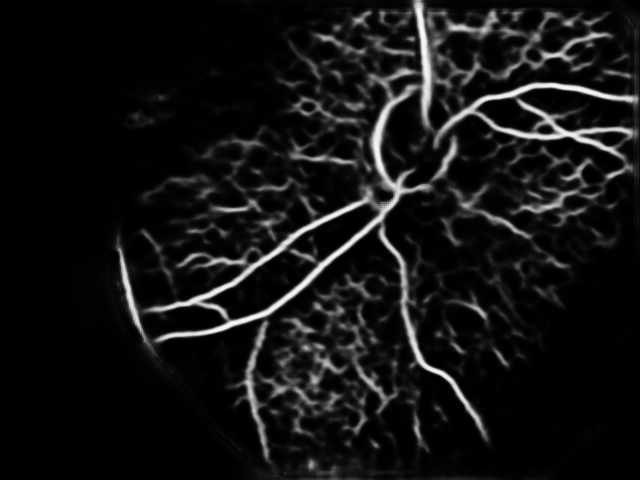}
    \end{minipage}
    \hfill
    \begin{minipage}{0.243\textwidth}
        \centering
        \includegraphics[width=\textwidth]{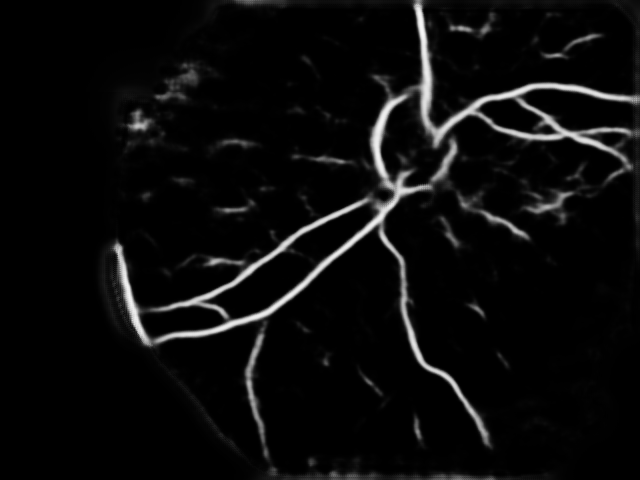}
    \end{minipage}
    
    \begin{minipage}{0.243\textwidth}
        \centering
        \includegraphics[width=\textwidth]{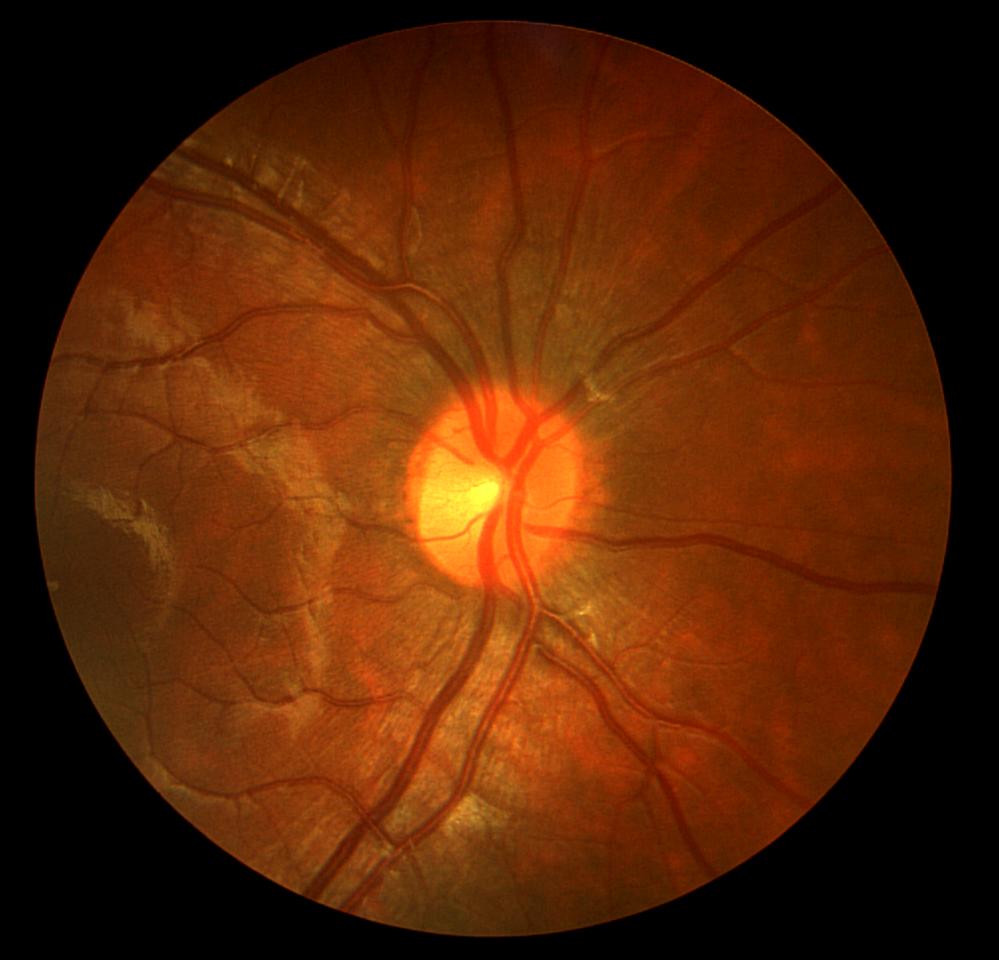}
        (a)
    \end{minipage}
    \hfill
    \begin{minipage}{0.243\textwidth}
        \centering
        \includegraphics[width=\textwidth]{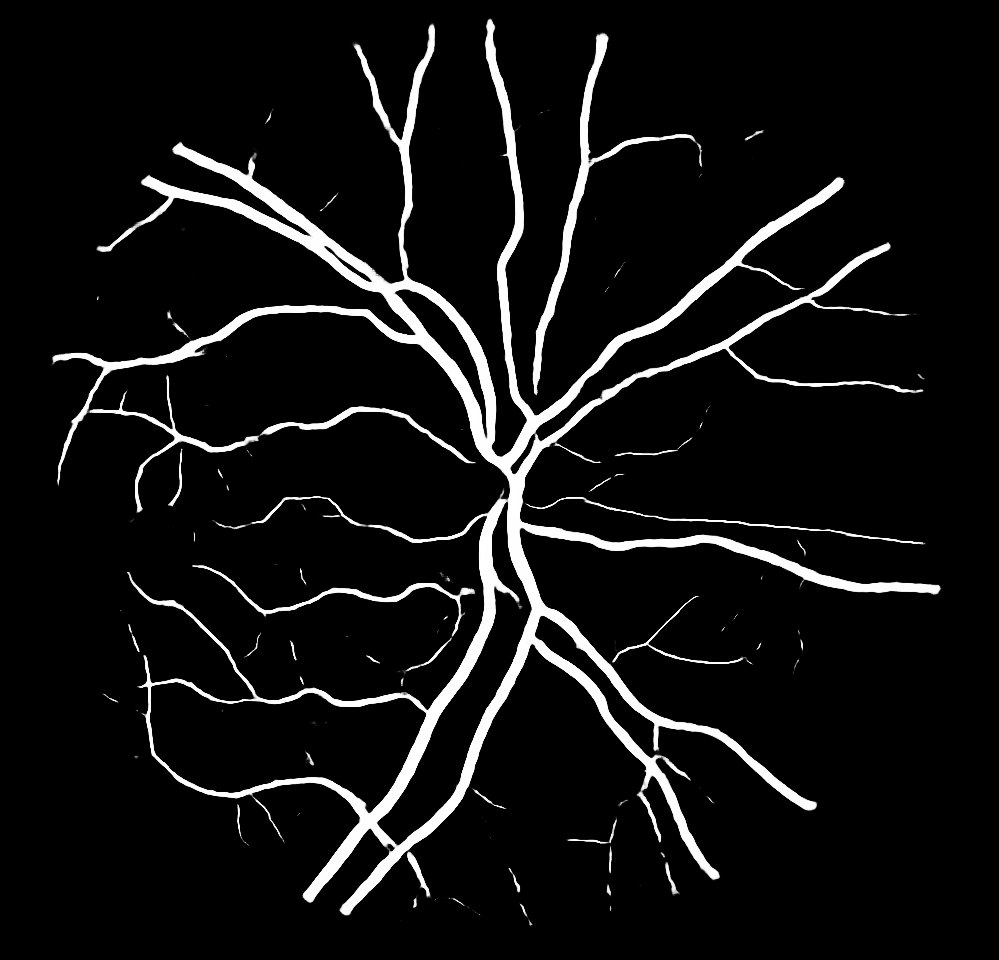}
        (b)
    \end{minipage}
    \hfill
    \begin{minipage}{0.243\textwidth}
        \centering
        \includegraphics[width=\textwidth]{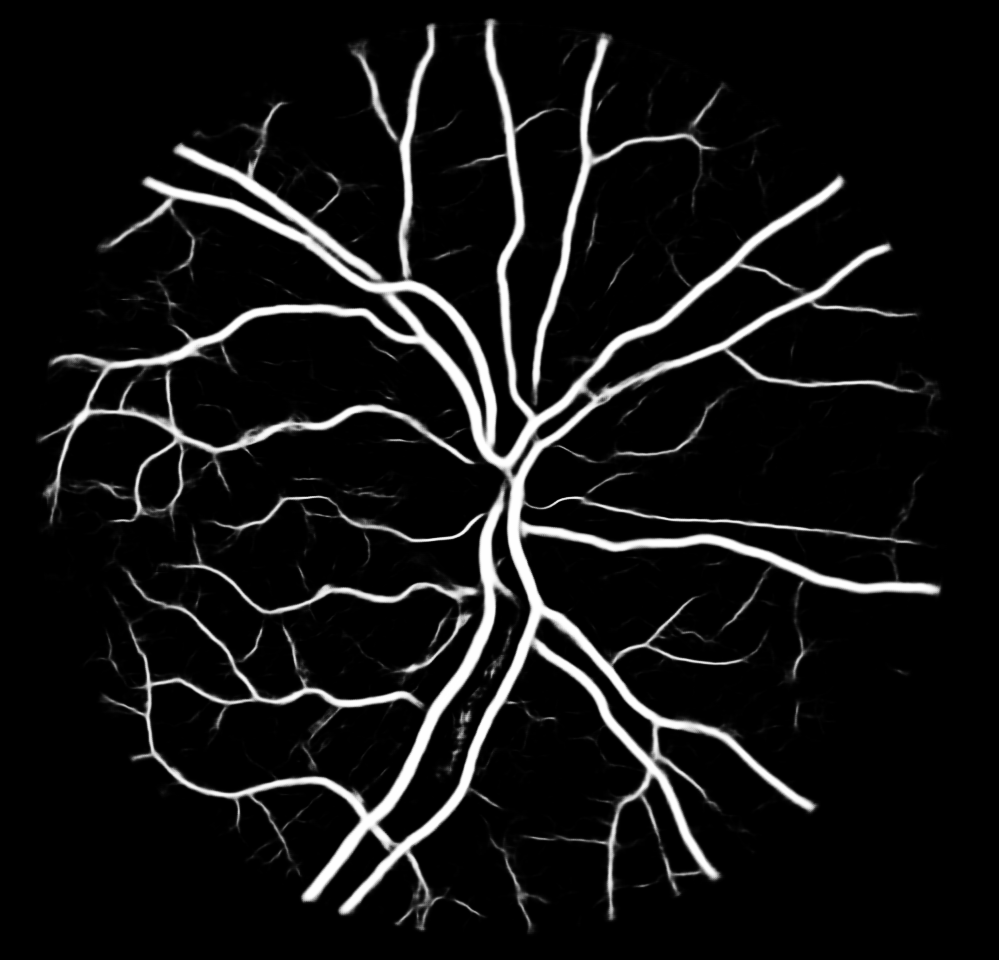}
        (c)
    \end{minipage}
    \hfill
    \begin{minipage}{0.243\textwidth}
        \centering
        \includegraphics[width=\textwidth]{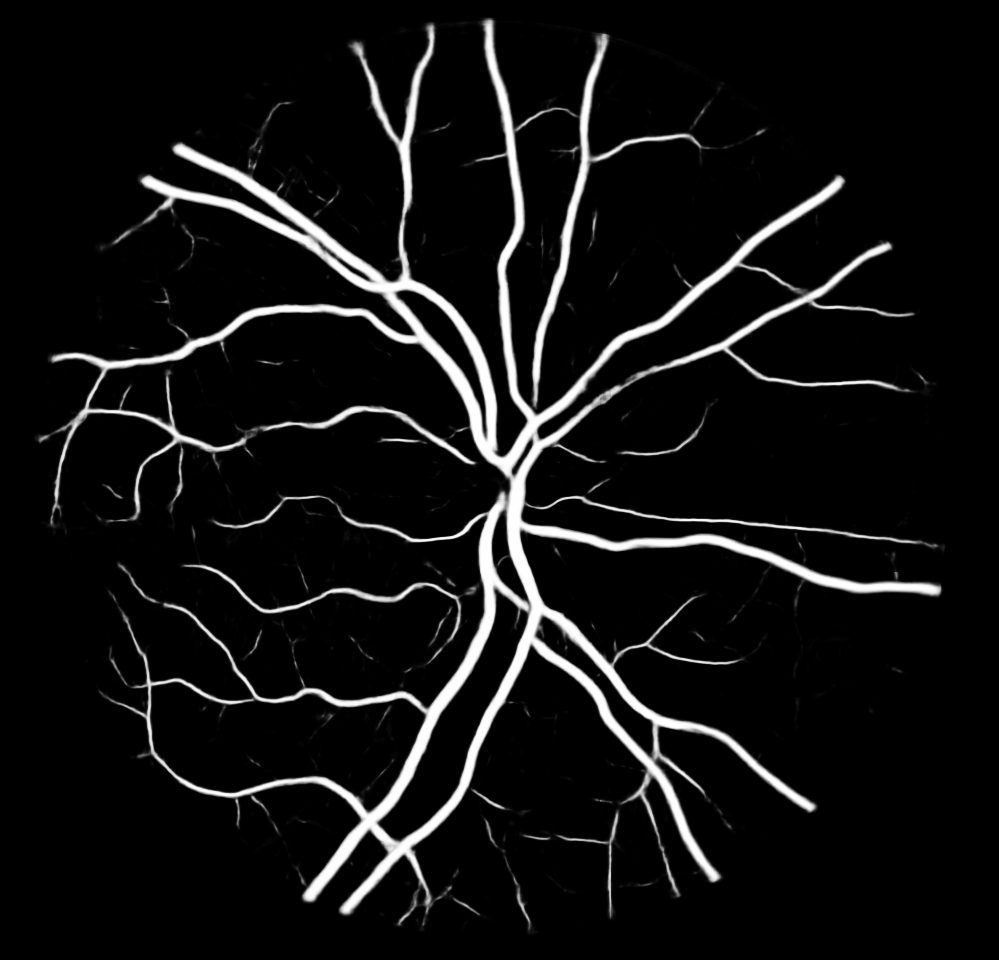}
        (d)
    \end{minipage}
    
    \caption{\textbf{Likelihoods maps generated by U-net with different weights variations:} The rows correspond to a sample from the DRIVE, STARE, AV-WIDE, VEVIO (Mosaics), VEVIO (Frames), and CHASE-DB datasets, resp. \textbf{(a)} Original image. \textbf{(b)} Likelihood map when trained with class weights estimated from the training data. \textbf{(c)} Likelihood map when using equal weights for both classes. \textbf{(d)} Likelihood map when using dynamic weights (see text for details). Using dynamic weights allows the network to capture more information about the vascular structure without introducing additional noise. Figure best viewed on-screen.}
    \label{fig:pmap_datasets}
\end{figure}

Figure~\ref{fig:pmap_datasets} contrast the outputs of our stochastic U-net vs. two fixed-weight U-nets on the datasets used in our experiments (see figure for details). Overall, networks trained with fixed weights struggle to capture faint vessels, as well as the pixels at the edges of thicker vessels. Interestingly, our stochastic weights are also better able to ignore vessel-like artifacts, e.g., the rim of a lens as seen in Fig.~\ref{fig:pmap_datasets}.








\begin{figure}[t]
\includegraphics[width=0.99\textwidth]{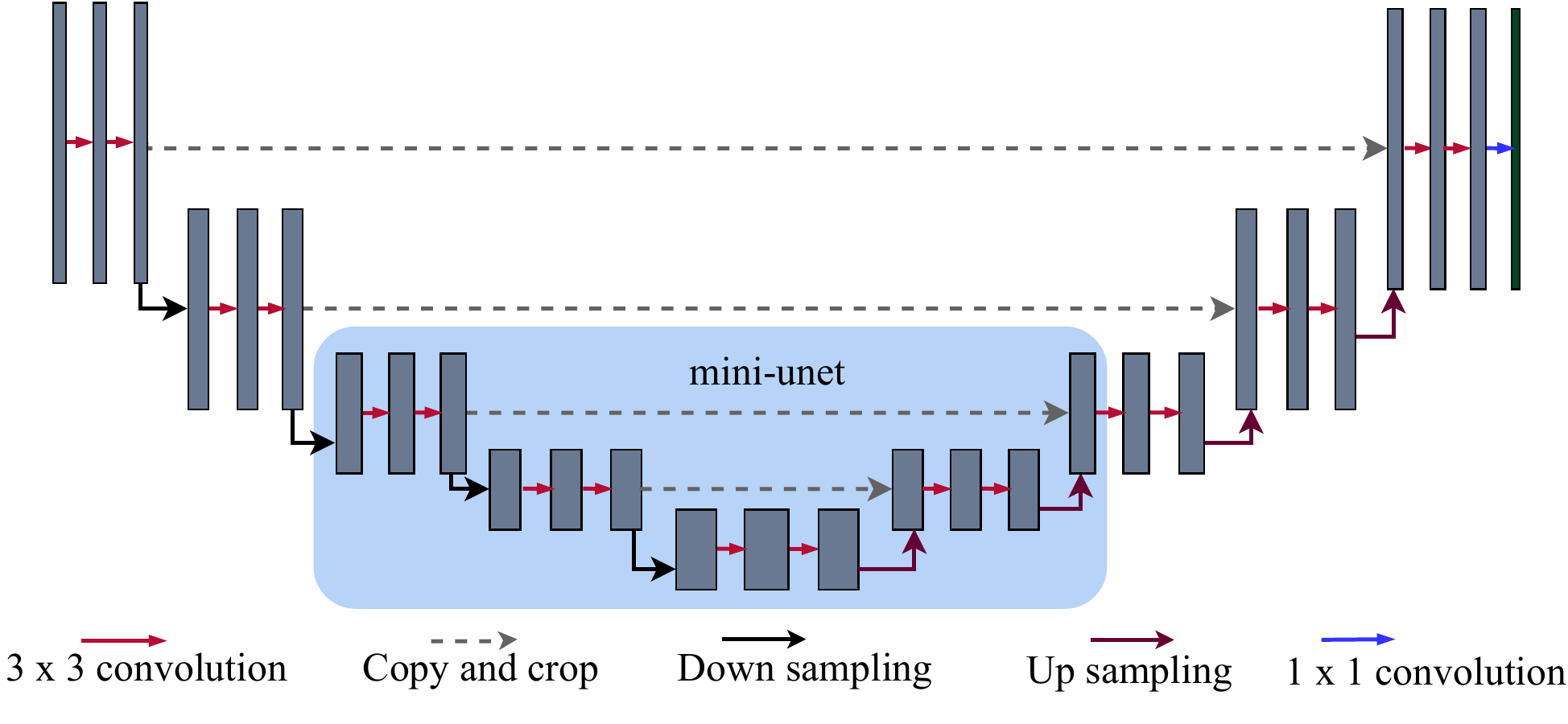}
\caption{\textbf{U-net and mini-U-net architectures:} As originally detailed in \cite{Ronneberger2015UNetCN}, the first half of the U-net architecture is a standard sequence of convolution and max-pooling layers. The layers in the second half, on the other hand, receive two inputs: the features from the previous layer and the features from the corresponding layer in the first half of the network (i.e., the one with the same dimensionality). The features from previous layers are upsampled to match the larger dimensions of subsequent layers. Intuitively, the upsampled features provide more global information, while the features transferred from the earlier layer give more local information. Our mini-U-net architecture (highlighted by the light blue square) consists of the middle layers of the full U-net.}
\label{unet_arch}
\end{figure}

\subsection{Support-resistance segmentation (SRS) for easy pixels}
The likelihood map generated in the previous step ranges from [0,1]. Pixels with a higher probability of being vessels have values closer to one (and closer to zero otherwise). Intuitively, the closer a value is to either extreme, the more confident the network is in its classification. Thus, we use this likelihood map to separate pixels into either \textit{easy} or \textit{ambiguous} classes. The top and bottom of Fig.~\ref{flow_ature}(c) show the easy and ambiguous pixels, respectively, of the likelihood map in Fig.~\ref{flow_ature}(b).

We use two thresholds to classify pixels into these two classes. We refer to the lower threshold as \textit{support} and the upper threshold as \textit{resistance}. All pixels between \textit{resistance} and \textit{support} are considered ambiguous. Pixels below \textit{support} (those with values near zero), which account for the majority of pixels, are background pixels whereas pixels above the \textit{resistance} (those with value near one) are considered vessel pixels.  

The choice of \textit{support} and \textit{resistance} thresholds affects the trade-off between precision and recall. Thresholds close to one and zero, respectively, lead to higher precision for easy pixels but cause more pixels to fall in the intermediate, ambiguous band. Thresholds closer to the middle lead to higher recall, but affect precision. For our experiments, we empirically set a \textit{support} of 20 and a \textit{resistance} of 235. 

\subsection{Targeted prediction for ambiguous pixels}
The last stage of our pipeline segments ambiguous pixels, i.e., those that fall within the \textit{support-resistance} band described above. Figs.~\ref{flow_ature}(d)-(g) illustrate the various steps involved in this targeted prediction. Generally, most vessel pixels will lie above the support threshold, so they have already been predicted with high precision. To classify the pixels with intermediate values, we train a second, patch-based network only on those pixels. As Fig.~\ref{unet_arch} shows, this mini-U-net network is a scaled-down version that uses only the middle layers of the full U-net architecture. We use two channels as inputs to this network: the full likelihood map outputted by the first U-net (Fig.~\ref{flow_ature}(b)) and a version of this likelihood map which only contains ambiguous pixels (the bottom image in Fig.~\ref{flow_ature}(c)).


Unlike prior patch-based approaches (e.g., Ciresan \textit{et al.} \cite{NIPS2012_4741}), we do not extract a patch for every pixel. Instead, we use a sparse set of $p\times p$ patches centered at the intersections between ambiguous pixels and a regular lattice, as described below:
\begin{enumerate}
    \item \textbf{Raw estimate:} We first produce a high-recall binary mask by considering all pixels above the support threshold as vessels. Intuitively, this mask should contain almost no false negatives, but will have numerous false positives. We then skeletonize the largest connected component in this mask (Fig.~\ref{flow_ature}(d)).
    \item \textbf{Patch selection:} We then define a regular lattice of width $p/2$ (Fig.~\ref{flow_ature}(e)) and determine the intersection points between this lattice and the skeleton from the previous step (Fig.~\ref{flow_ature}(d)). We empirically verified that the union of the patches defined by these intersections always covered every ambiguous pixel.
    \item \textbf{Mini-U-net classification:} Finally, we classify the ambiguous pixels in each patch using our second network (Fig.~\ref{flow_ature}(g)). For each patch, we extract the corresponding regions in the full likelihood map and the map with only ambiguous pixels, resp. Intuitively, the second channel indicates which pixels are most crucial to classify correctly. Finally, we use the outputs from the second network as the final labels for ambiguous pixels (Fig.~\ref{flow_ature}(h)).
\end{enumerate}

\section{Experiments and Results}
\label{sec:experiments}
We carried out experiments on five different retinal vessel datasets to verify the effectiveness of our proposed approach. Below, we first detail our experimental protocol and then discuss our results.




\subsection{Materials and methods}

\noindent{\textbf{Hardware:}} All our experiments were conducted on a Dell Precision 7920R server with two Intel Xeon Silver 4110 CPUs, two GeForce GTX 1080 Ti graphics cards, and 128 GBs of RAM.

\vspace{0.75em}
\noindent \textbf{Datasets:} We used five datasets for our experiments (see the first column of Fig.~\ref{fig:pmap_datasets} for a sample image from each dataset):
\begin{itemize}
    \item DRIVE \cite{drive_dataset}: 40, 565$\times$584 color fundus images centered on the macula. Each image includes a circular field-of-view (FOV) mask. This dataset is pre-divided into 20 training and 20 test images, but we further divided the training set randomly into 15 training and 5 validation images.
    \item STARE \cite{845178}: 20, 700$\times$ 605 color images with no FOV masks, also centered on the macula. Many images exhibit some degree of pathology (e.g., exudates).
    \item AV-WIDE \cite{wide-dataset:6987362,7120990}: 30 wide-FOV images. Image sizes vary, but are around 1300$\times$800 pixels for most images. Images are loosely centered on the macula. As explained below, we used the existing, graph-based annotations to manually segment these images.
    \item VEVIO \cite{vevio-dataset:2012}: This dataset has two types of images: 16, 640$\times$480 frames from 16 different video indirect ophthalmoscopy (VIO) videos and 16 corresponding mosaics (around 600$\times$500 pixels) obtained by fusing different frames into a single image \cite{Estrada:VIO-mosaicing}. Due to the difficult image acquisition process, these images are blurry and have lens artifacts.
    \item CHASE-DB \cite{6224174}: 14 left-to-right pairs of color images centered on the optic nerve (28 images total). Images are 999$\times$960 pixels.
\end{itemize}

\vspace{0.75em}
\noindent \textbf{Data preparation:} We used only the green channel of each image. We then applied contrast adaptive histogram equalization \cite{Zuiderveld:1994:CLA:180895.180940} as a preprocessing step to enhance contrast. Given the limited number of images in each dataset, we used image augmentation techniques to increase our data size. Specifically, we flipped the input image horizontally and vertically randomly in each iteration.

\vspace{0.75em}
\noindent \textbf{Ground truth preparation:} All datasets except AV-WIDE had preexisting ground-truth pixel labels. The original AV-WIDE dataset had only graph-based labels, so we manually traced the vessels in Adobe Photoshop CS (Adobe Systems Inc., San Jose, CA) using a Wacom Intuous3 graphics tablet (Wacom Co. Ltd, Kazo-shi, Saitama, Japan) to obtain pixel-level segmentations. The original, graph-based annotations had been carried out by an expert ophthalmologist, so we followed them exactly. 

\vspace{0.75em}
\noindent \textbf{Network architectures:} As noted above, we used a full-sized U-net architecture for our initial probability map and a smaller, mini-U-net network for targeted prediction. The U-net architecture consist of a series of down-sampling steps followed by the same number of up-sampling steps. Each up-sampling step receives, crops to match the size, and concatenates the feature map from the down-sample step on the same level, as shown in Fig.~\ref{unet_arch}. We used 3$ \times $3-pixel kernels except in the last layer; here, we used 1$ \times $1 kernels to produce two outputs, which are then passed through a \textit{softmax} layer to obtain two probabilities: one for that particular pixel being a vessel and another for it being part of the background. Depending on which of the probabilities is higher, a pixel is labeled as either \textit{vessel} or \textit{background}. As in the original U-net architecture \cite{Ronneberger2015UNetCN}, this network receives inputs of size 572$ \times $572 pixels, but only predicts values for the internal 388$ \times $388 pixels. The outer band of 92$ \times $92 pixel allows the U-net a wider view of the vascular structure. As in \cite{Ronneberger2015UNetCN}, we shift the U-net to obtain a label for every pixel and mirror the 388$\times$ 388 region if the outer band extends beyond the image. 

Our second network, the mini-U-net architecture, is identical to the middle three layers of the full-size U-net (see Fig.~\ref{unet_arch}). Thus, its input size is 140 $\times$ 140 and it produces labels for the middle 100$\times$100 pixels. Similar to the first network, we expand and mirror patches as needed. However, we train this network using a \textit{dice loss}, rather than cross entropy because our goal is to maximize the F1 score of our pipeline. The weighted dice loss is given by the following formula:
\begin{equation}
	\label{eq:dice_loss}
	F_{\beta} = (1 + \beta^{2}) \cdot \frac{precision \cdot recall}{\beta^2 \cdot precision + recall}
\end{equation}, 
where $\beta$ controls how much we want to favor precision vs. recall. Higher beta yields higher precision and vice versa. As with the full-size U-net, we used a stochastic loss in which we sampled $\beta$ for each training iteration:
\begin{equation}
	\label{eq:dyn_dice_loss}
	F_{\beta} = (1 + B_{rand}(1, \alpha, s)^{2}) \cdot \frac{precision \cdot recall}{B_{rand}(1, \alpha, s)^{2} \cdot precision + recall}
\end{equation},
where $B_{rand}(1, \alpha, s)$ is picked randomly within range $1$ - $\alpha$ with a stepsize of $s$.

\vspace{0.75em}
\noindent \textbf{Training parameters:} We trained our networks using Adam optimization \cite{2014arXiv1412.6980K} with a learning rate of 0.001 and a mini-batch size of 4. We trained the full U-net for 250 epochs and the mini-U-net for 60 epochs. For the full U-net, we used stochastic weights of $w_{rand}(1, 100, 1)$, i.e., where weights randomly oscillated within a range of $1-100$ with a step size of $1$. For the mini-U-net, we used a dice loss with stochastic weights of $B_{rand}$(1, 2, 0.1), where $\beta$ randomly oscillated between $1-2$ with a step size of $0.1$.

\vspace{0.75em}
\noindent \textbf{Model validation:} We used 5-fold cross validation with training, validation, and test sets on all datasets except DRIVE, since this dataset already had a predefined test set. For each fold, we randomly split the training set into 75\% training and 25\% validation images. In our tables, we report the performance on the test set across all the folds. Note that, using this protocol, each image is only included once in a test set.

\subsection{Pipelines tested}
To better understand the impact of each stage of our pipeline, we tested two versions of our approach, as well as two baseline methods:
\begin{itemize}
  \item \textit{Dynamic weights with targeted prediction:} Our proposed approach.
  \item \textit{Dynamic weights only:} We trained the full U-net using dynamic class weights, but omitted the targeted prediction step.
  \item \textit{Fixed weights only:} We trained the full U-net using fixed class weights based on the ratio of vessel to background pixels in the training set. No targeted prediction step.
  \item \textit{Fixed weights with targeted prediction:} We applied both fixed weights and targeted prediction.
\end{itemize}
The parameters for the Mini-U-net at the end of the full pipelines (variations 1 and 4) were the same for both configurations. In total, we carried out twenty experiments across the five datasets.

\subsection{Results}
We quantified our method's performance in terms of \textit{precision}, \textit{recall} and \textit{F1 score} (also called F-measure). For comparison to the state of the art, we also calculated the accuracy of our model, even though this value can be misleading for unbalanced datasets \cite{vevio-dataset:2012}. The precision, recall, and accuracy values in our tables are the mean values across all test images. The reported F1 score is the harmonic average of the corresponding mean precision and recall values. As Forman \textit{et al.} showed \cite{Forman:2010:ACS:1882471.1882479}, if the positive cases constitute around 7\% or more of the samples, then this is an unbiased way to calculate the mean F1 score for $k$-fold cross validation. Across the different datasets, vessel pixels constituted around 10\% of all pixels.



\begin{table}[h]
\centering
\caption{Results on DRIVE dataset}
\begin{small}
    \begin{tabular}{|l|c|c|c|c|}
        \hline
        \multicolumn{1}{|c|}{\bfseries Method} & \multicolumn{1}{c|}{\bfseries Precision} & \multicolumn{1}{c|}{\bfseries Recall} & \multicolumn{1}{c|}{\bfseries F$_1$-Score} &
        \multicolumn{1}{c|}{\bfseries Accuracy}\\ \hline
         \multicolumn{1}{|p{4cm}|}{\raggedright U-net\cite{r2unet:journals/corr/abs-1802-06955}} & 0.8852 & 0.7537 & 0.8142 & 0.9531 \\ \hline
        \multicolumn{1}{|p{4cm}|}{\raggedright Residual U-net\cite{r2unet:journals/corr/abs-1802-06955}} & \textbf{0.8614} & 0.7726 & 0.8149 & 0.9553 \\ \hline
        \multicolumn{1}{|p{4cm}|}{\raggedright Recurrent U-net\cite{r2unet:journals/corr/abs-1802-06955}} & 0.8603 & 0.7751 & 0.8155 & 0.9556 \\ \hline
        \multicolumn{1}{|p{4cm}|}{\raggedright R2 U-net\cite{r2unet:journals/corr/abs-1802-06955}} & 0.8589 & 0.7792 & 0.8171 & 0.9556 \\ \hline
        \multicolumn{1}{|p{4cm}|}{\raggedright Conditional GAN \cite{con_gan_unet:journals/corr/abs-1805-04224}} & 0.8143 & 0.8274 & 0.8208 & 0.9608 \\ \hline
        \multicolumn{1}{|p{4cm}|}{\raggedright LadderNet\cite{laddernet:journals/corr/abs-1810-07810}} & 0.8593 & 0.7856 & 0.8208 & 0.9561 \\ \hline
        \multicolumn{1}{|p{4cm}|}{\raggedright DUNet\cite{2018arXiv181101206J}} & 0.8537 & 0.7894 & 0.8203 & \textbf{0.9697} \\ \hline
        \multicolumn{1}{|p{4cm}|}{\raggedright\bfseries Fixed weights only} & 0.7657 & \bfseries{0.8410} & 0.8015 & 0.9633  \\ \hline
        \multicolumn{1}{|p{4cm}|}{\raggedright\bfseries  Fixed weights with targeted prediction} & 0.7823 & 0.8246 & 0.8028 & 0.9643 \\ \hline
        \multicolumn{1}{|p{4cm}|}{\raggedright\bfseries Dynamic weights only} & 0.8323 & 0.8163 & 0.8242 & 0.9692 \\ \hline
        \multicolumn{1}{|p{4cm}|}{\raggedright\bfseries Dynamic weights with targeted prediction} & 0.8284 & 0.8235 & {\bfseries 0.8259} & 0.9693 \\ \hline
    \end{tabular}
\end{small}
    \label{tab:drive}
\end{table}

Tables~\ref{tab:drive} to \ref{tab:vevioFrames} summarize our results on the various datasets. A value in bold represents the best score for that particular column. For convenience, we also included those of current state-of-the-art approaches. Note, though, that we only list methods that reported F1 scores (or, equivalently, both precision and recall values). Due to the strong class imbalance between the two classes, accuracy alone is not an informative measure of performance. As the various tables show, our proposed pipeline consistently outperformed existing methods. To the best of our knowledge, our F1 scores and most of our accuracy values are state-of-the-art results across the five datasets. 

Overall, dynamic weights outperformed fixed weights across the various datasets, including the highly studied DRIVE and STARE datasets. Specifically, stochastic weights led to more balanced precision and recall scores. For example, in the DRIVE dataset, the full, stochastic pipeline obtained a precision of 0.8284, and a recall of 0.8235 while the fixed-weight pipeline gave a more unbalanced results (precision of 0.7823 and recall of 0.8246). This pattern was replicated across datasets. Compared to prior methods, we consistently achieved better recall with little or no sacrifice in precision. Existing methods with high precision and low recall seldom misclassify a background pixel as a vessel, but they struggle to detect faint vessels.

Notably, our method outperformed other extensions of the original U-net architecture, including LadderNet ~\cite{laddernet:journals/corr/abs-1810-07810}, R2Unet ~\cite{r2unet:journals/corr/abs-1802-06955}, and DUNet \cite{2018arXiv181101206J}. The difference is particularly large for the CHASE-DB dataset (Table~\ref{tab:CHASE-DB}). Here, our method $F_{1}$ score was $2.66\%$ higher than the previous state of the art (LadderNet). 


\begin{table}[h]
\centering
\caption{Results on STARE dataset}
\begin{small}
    \begin{tabular}{|l|c|c|c|c|}
        \hline
        \multicolumn{1}{|c|}{\bfseries Method} & \multicolumn{1}{c|}{\bfseries Precision} & \multicolumn{1}{c|}{\bfseries Recall} & \multicolumn{1}{c|}{\bfseries F$_1$-Score} &
        \multicolumn{1}{c|}{\bfseries Accuracy}\\ \hline
        \multicolumn{1}{|p{4cm}|}{\raggedright U-net\cite{r2unet:journals/corr/abs-1802-06955}} & 0.8475 & 0.8270 & 0.8373 & 0.9690 \\ \hline
        \multicolumn{1}{|p{4cm}|}{\raggedright Residual U-net\cite{r2unet:journals/corr/abs-1802-06955}} & 0.8581 & 0.8203 & 0.8388 & 0.9700 \\ \hline
        \multicolumn{1}{|p{4cm}|}{\raggedright Recurrent U-net\cite{r2unet:journals/corr/abs-1802-06955}} & \textbf{0.8705} & 0.8108 & 0.8396 & 0.9706 \\ \hline
        \multicolumn{1}{|p{4cm}|}{\raggedright R2 U-net\cite{r2unet:journals/corr/abs-1802-06955}} & 0.8659 & 0.8298 & 0.8475 & 0.9712 \\ \hline
        \multicolumn{1}{|p{4cm}|}{\raggedright Conditional GAN \cite{con_gan_unet:journals/corr/abs-1805-04224}} & 0.8466 & 0.8538 & 0.8502 & 0.9771 \\ \hline
        \multicolumn{1}{|p{4cm}|}{\raggedright DUNet\cite{2018arXiv181101206J}} & 0.8856 & 0.7428 & 0.8079 & 0.9729 \\ \hline
         \multicolumn{1}{|p{4cm}|}{\raggedright\bfseries Fixed weights only} & 0.8138 & 0.8538 & 0.8480 & 0.9739  \\ \hline
        \multicolumn{1}{|p{4cm}|}{\raggedright\bfseries Fixed weights with targeted prediction} & 0.7979 & \bfseries{0.8692} & 0.8320 & 0.9732 \\ \hline
        \multicolumn{1}{|p{4cm}|}{\raggedright\bfseries Dynamic weights only} & 0.8413 &  0.8424 & 0.8418 &  0.9758 \\ \hline
        \multicolumn{1}{|p{4cm}|}{\raggedright\bfseries Dynamic weights with targeted prediction} & 0.8559 & 0.8541 & {\bfseries 0.8549} & {\bfseries 0.9780} \\ \hline
    \end{tabular}
\end{small}
    \label{tab:stare}
\end{table}



\begin{table}[h]
\centering
\caption{Results on CHASE-DB dataset}
\begin{small}
    \begin{tabular}{|l|c|c|c|c|}
        \hline
        \multicolumn{1}{|c|}{\bfseries Method} & \multicolumn{1}{c|}{\bfseries Precision} & \multicolumn{1}{c|}{\bfseries Recall} & \multicolumn{1}{c|}{\bfseries F$_1$-Score} &
        \multicolumn{1}{c|}{\bfseries Accuracy}\\ \hline
        \multicolumn{1}{|p{4cm}|}{\raggedright U-net\cite{r2unet:journals/corr/abs-1802-06955}} & 0.7336 & 0.8288 & 0.7783 & 0.9578 \\ \hline
        \multicolumn{1}{|p{4cm}|}{\raggedright Residual U-net\cite{r2unet:journals/corr/abs-1802-06955}} & 0.7857 & 0.7726 & 0.7800 & 0.9553 \\ \hline
        \multicolumn{1}{|p{4cm}|}{\raggedright Recurrent U-net\cite{r2unet:journals/corr/abs-1802-06955}} & 0.8195 & 0.7459  & 0.7810 & 0.9622 \\ \hline
        \multicolumn{1}{|p{4cm}|}{\raggedright R2 U-net\cite{r2unet:journals/corr/abs-1802-06955}} & 0.81072 & 0.0.7756 & 0.7928 & 0.9634 \\ \hline
        \multicolumn{1}{|p{4cm}|}{\raggedright LadderNet \cite{laddernet:journals/corr/abs-1810-07810}} & 0.8084 & 0.7978 & 0.8031 & 0.9656 \\ \hline
        \multicolumn{1}{|p{4cm}|}{\raggedright DUNet\cite{2018arXiv181101206J}} & 0.7510 & 0.8229 & 0.7853 & 0.9724 \\ \hline
         \multicolumn{1}{|p{4cm}|}{\raggedright\bfseries Fixed weights only} & 0.8089 & 0.8271 & 0.8179 & 0.9744  \\ \hline
        \multicolumn{1}{|p{4cm}|}{\raggedright\bfseries Fixed weights with targeted prediction} & 0.8266 & 0.8085 & 0.8174 & 0.9749 \\ \hline
        \multicolumn{1}{|p{4cm}|}{\raggedright\bfseries Dynamic weights only} & 0.8175 &  \bfseries{0.8296} & 0.8235 &  0.9753 \\ \hline
        \multicolumn{1}{|p{4cm}|}{\raggedright\bfseries Dynamic weights with targeted prediction} & \bfseries{0.8550} &  0.8143 & \bfseries{0.8245} & {\bfseries 0.9759} \\ \hline
    \end{tabular}
\end{small}
    \label{tab:CHASE-DB}
\end{table}

\begin{table}[h]
\centering
    \caption{Results on AV-WIDE dataset}
\begin{small}
    \begin{tabular}{|l|c|c|c|c|}
        \hline
        \multicolumn{1}{|c|}{\bfseries Method} & \multicolumn{1}{c|}{\bfseries Precision} & \multicolumn{1}{c|}{\bfseries Recall} & \multicolumn{1}{c|}{\bfseries F$_1$-Score} &
        \multicolumn{1}{c|}{\bfseries Accuracy}\\ \hline
        \multicolumn{1}{|p{4cm}|}{\raggedright\bfseries Fixed weights only} & 0.7611 & 0.7898 & 0.7751 & 0.9706  \\ \hline
        \multicolumn{1}{|p{4cm}|}{\raggedright\bfseries Fixed weights with targeted prediction} & 0.7439 & {\bfseries 0.8083} & 0.7747 & 0.9698 \\ \hline
        \multicolumn{1}{|p{4cm}|}{\raggedright\bfseries Dynamic weights only} & 0.8154 & 0.7751 & 0.7947 & 0.9742 \\ \hline
        \multicolumn{1}{|p{4cm}|}{\raggedright\bfseries Dynamic weights with targeted prediction} & {\bfseries 0.8283} &  0.7815 & {\bfseries 0.8042} & {\bfseries 0.9755} \\ \hline
    \end{tabular}
\end{small}    
    \label{tab:AV-WIDE}
\end{table}

Table~~\ref{tab:AV-WIDE} shows our results for the AV-WIDE dataset. Our paper is the first assessment of this dataset as we manually created the ground truth for it. As with the other datasets, we can observe a similar pattern of unbalanced precision and recall for fixed class weights. In contrast, our stochastic technique has more balanced precision and recall. Moreover, the $F_1$ score and accuracy are significantly higher for our method, particularly for the full pipeline with targeted prediction.

\begin{table}[h]
\centering
\caption{Results on VEVIO dataset (Mosaics)}
\begin{small}
    \begin{tabular}{|l|c|c|c|c|}
        \hline
        \multicolumn{1}{|c|}{\bfseries Method} & \multicolumn{1}{c|}{\bfseries Precision} & \multicolumn{1}{c|}{\bfseries Recall} & \multicolumn{1}{c|}{\bfseries F$_1$-Score} &
        \multicolumn{1}{c|}{\bfseries Accuracy}\\ \hline
        \multicolumn{1}{|p{4cm}|}{\raggedright Forest-based Dijkstra \cite{vevio-dataset:2012}} & - & - & 0.5053 & 0.9573  \\ \hline        \multicolumn{1}{|p{4cm}|}{\raggedright\bfseries Fixed weights only} & 0.5537 & 0.6832 & 0.6093 & 0.9637  \\ \hline
        \multicolumn{1}{|p{4cm}|}{\raggedright\bfseries Fixed weights with targeted prediction} & 0.5472 & \bfseries{0.6984} & 0.6136 & 0.9632 \\ \hline
        \multicolumn{1}{|p{4cm}|}{\raggedright\bfseries Dynamic weights only} & 0.6147 & 0.6355 & 0.6249 & 0.9683 \\ \hline
        \multicolumn{1}{|p{4cm}|}{\raggedright\bfseries Dynamic weights with targeted prediction} & {\bfseries 0.6573} & 0.6739 & {\bfseries 0.6654} & {\bfseries 0.9719} \\ \hline
    \end{tabular}
\end{small}
    \label{tab:vevioMosaics}
\end{table}

\begin{table}[h]
\centering
\caption{Results on VEVIO dataset (Frames)}
\begin{small}
    \begin{tabular}{|l|c|c|c|c|}
        \hline
        \multicolumn{1}{|c|}{\bfseries Method} & \multicolumn{1}{c|}{\bfseries Precision} & \multicolumn{1}{c|}{\bfseries Recall} & \multicolumn{1}{c|}{\bfseries F$_1$-Score} &
        \multicolumn{1}{c|}{\bfseries Accuracy}\\ \hline
        \multicolumn{1}{|p{4cm}|}{\raggedright Forest-based Dijkstra \cite{vevio-dataset:2012}} & - & - & 0.5403 & 0.9101  \\ \hline        \multicolumn{1}{|p{4cm}|}{\raggedright\bfseries Fixed weights only} & 0.5212 & \bfseries{0.6580} & 0.5817 & 0.9569  \\ \hline
        \multicolumn{1}{|p{4cm}|}{\raggedright\bfseries Fixed weights with targeted prediction} & 0.5328 & 0.6482 & 0.5849 & 0.9581 \\ \hline
        \multicolumn{1}{|p{4cm}|}{\raggedright\bfseries Dynamic weights only} & 0.5924 & 0.5896 & \bfseries{0.5910} & \bfseries{0.9629} \\ \hline
        \multicolumn{1}{|p{4cm}|}{\raggedright\bfseries Dynamic weights with targeted prediction} & {\bfseries 0.60052} & 0.5435 & 0.5706 & 0.9628 \\ \hline
    \end{tabular}
\end{small}
    \label{tab:vevioFrames}
\end{table}

Finally, our method also achieved state-of-the-art results in the highly challenging VEVIO dataset, which has two sets of images: composite mosaics and individual frames taken from a low-resolution video \cite{Estrada:VIO-mosaicing}. As with the other datasets, fixed class weights gave highly biased scores in terms of precision and recall, while dynamic weights gave more balanced results (Tables~\ref{tab:vevioMosaics} and \ref{tab:vevioFrames}).

\begin{table}[h]
\centering
\caption{Equal weights \{1,1\} vs. class-based weights across datasets. Results shown are for U-net without targeted prediction.}
\begin{small}
    \begin{tabular}{|l|c|c|c|c|}
        \hline
        \multicolumn{1}{|c|}{\bfseries Dataset and class weights} & \multicolumn{1}{c|}{\bfseries Precision} & \multicolumn{1}{c|}{\bfseries Recall} & \multicolumn{1}{c|}{\bfseries F$_1$-Score} &
        \multicolumn{1}{c|}{\bfseries Accuracy}\\ \hline
        \multicolumn{1}{|p{4cm}|}{\raggedright\bfseries DRIVE \{1, 1\}} & 0.8418 & 0.8023 & 0.8216 & 0.9694 \\ \hline
        \multicolumn{1}{|p{4cm}|}{\raggedright\bfseries DRIVE class-weighted}  & 0.7657 & 0.8410 & 0.8015 & 0.9633 \\ \hline
        \hline
        \multicolumn{1}{|p{4cm}|}{\raggedright\bfseries STARE \{1, 1\}} & 0.8559 & 0.8208 & 0.8379 &  0.9757 \\ \hline
        \multicolumn{1}{|p{4cm}|}{\raggedright\bfseries STARE class-weighted} & 0.8138 & 0.8538 & 0.8480 & 0.9739 \\ \hline
        \hline
         \multicolumn{1}{|p{4cm}|}{\raggedright\bfseries CHASE-DB \{1, 1\}} & 0.8332 & 0.8135 & 0.8232 &  0.9757 \\ \hline
         \multicolumn{1}{|p{4cm}|}{\raggedright\bfseries CHASE-DB class-weighted} & 0.8089 & 0.8271 & 0.8179 & 0.9744 \\ \hline
         \hline
        \multicolumn{1}{|p{4cm}|}{\raggedright\bfseries AV-WIDE \{1, 1\}} & 0.8231 & 0.7552 & 0.7876 &  0.9737 \\ \hline
        \multicolumn{1}{|p{4cm}|}{\raggedright\bfseries AV-WIDE class-weighted} & 0.7611 & 0.7898 & 0.7751 & 0.9706 \\ \hline
        \hline
        \multicolumn{1}{|p{4cm}|}{\raggedright\bfseries VEVIO Mosaic \{1, 1\}} & 0.6507 & 0.5564 & 0.5998 &  0.9690 \\ \hline
        \multicolumn{1}{|p{4cm}|}{\raggedright\bfseries VEVIO Mosaic class-weighted} & 0.5537 & 0.6832 & 0.6093 & 0.9637 \\ \hline
        \hline
        \multicolumn{1}{|p{4cm}|}{\raggedright\bfseries VEVIO Frame \{1, 1\}} & 0.6288 & 0.5257 & 0.5726 &  0.9643 \\ \hline
        \multicolumn{1}{|p{4cm}|}{\raggedright\bfseries VEVIO Frame class-weighted} & 0.5212 & 0.6580 & 0.5817 & 0.9569 \\ \hline
    \end{tabular}
\end{small}
    \label{tab:equal_weights}
\end{table}

\begin{table}[h]
\centering
\caption{Results of different combination of fixed weights in cross entropy loss function without targeted prediction for DRIVE dataset \{$w_0$, $w_1$\}=\{weight for class 0 or background pixels, and weight for class 1 or vessel pixel\}.}
\begin{small}
    \begin{tabular}{|l|c|c|c|c|}
        \hline
        \multicolumn{1}{|c|}{\bfseries Dataset and class weights} & \multicolumn{1}{c|}{\bfseries Precision} & \multicolumn{1}{c|}{\bfseries Recall} & \multicolumn{1}{c|}{\bfseries F$_1$-Score} &
        \multicolumn{1}{c|}{\bfseries Accuracy}\\ \hline
        \multicolumn{1}{|p{4cm}|}{\raggedright\bfseries DRIVE class-weighted (about \{1, 10\})} & 0.7657 & \bfseries{0.8410} & 0.8015 & 0.9633  \\ \hline
        \multicolumn{1}{|p{4cm}|}{\raggedright\bfseries DRIVE \{1, 5\}} & 0.7806 & 0.8264 & 0.8028 & 0.9642  \\ \hline
        \multicolumn{1}{|p{4cm}|}{\raggedright\bfseries DRIVE \{1, 1\}} & 0.8418 & 0.8023 & 0.8216 & 0.9694 \\ \hline
        \multicolumn{1}{|p{4cm}|}{\raggedright\bfseries DRIVE \{5, 1\}} & 0.8722 & 0.7270 & 0.7930 & 0.9665 \\ \hline
        \multicolumn{1}{|p{4cm}|}{\raggedright\bfseries DRIVE \{10, 1\}} & 0.8712 & 0.7137 & 0.7867 &  0.9654 \\ \hline
        \multicolumn{1}{|p{4cm}|}{\raggedright\bfseries DRIVE $w_{rand}(1, 10, 1)$} & \textbf{0.8508} & 0.7965 & 0.8227 &  \textbf{0.9696} \\ \hline
        \multicolumn{1}{|p{4cm}|}{\raggedright\bfseries DRIVE $w_{rand}(1, 100, 1)$ (our approach)} & 0.8323 & 0.8163 & \textbf{0.8242} & 0.9692 \\ \hline
    \end{tabular}
\end{small}
    
    \label{tab:different_weights_drive}
\end{table}

\begin{figure}[!t]
  \centering
    \begin{minipage}[b]{0.49\textwidth}
        \centering
        \includegraphics[width=\textwidth]{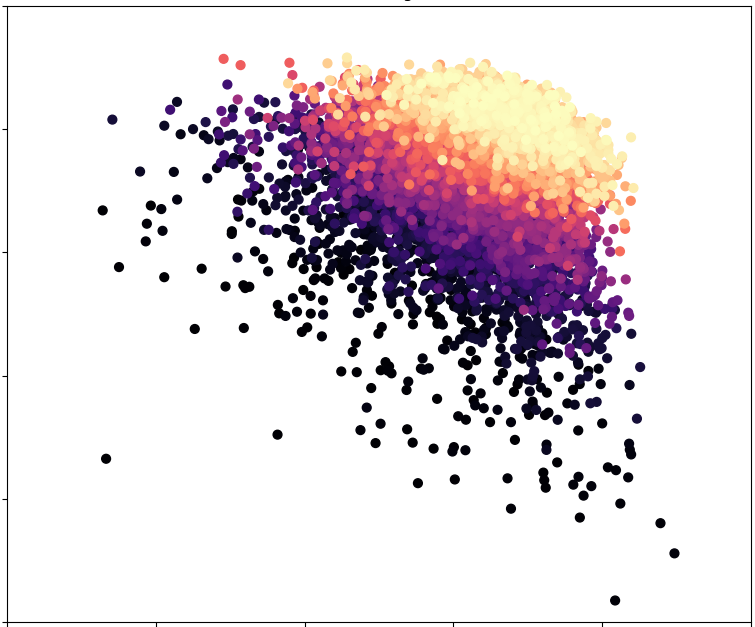}
        \label{fig:c}
        (a)
    \end{minipage}
    \begin{minipage}[b]{0.49\textwidth}
        \centering
        \includegraphics[width=\textwidth]{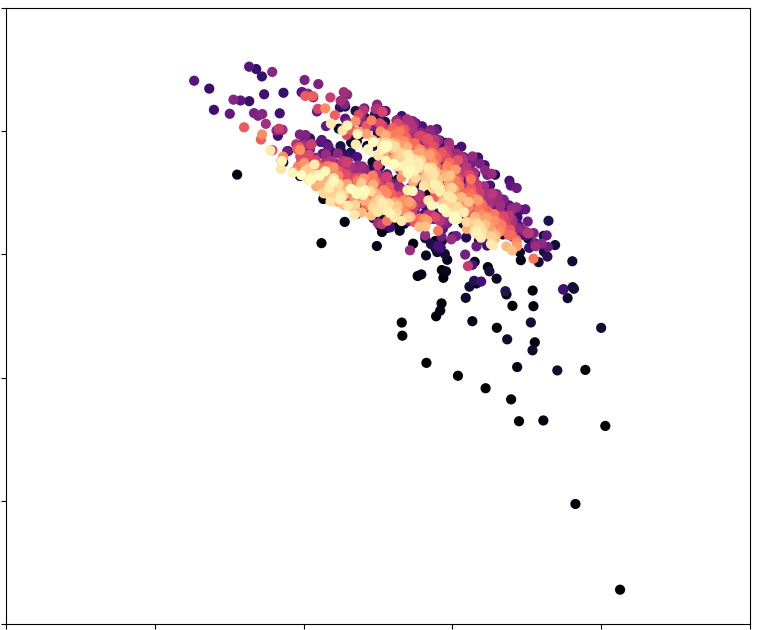}
        \label{fig:d}
        (b)
    \end{minipage}
    \begin{minipage}[b]{0.49\textwidth}
        \centering
        \includegraphics[width=\textwidth]{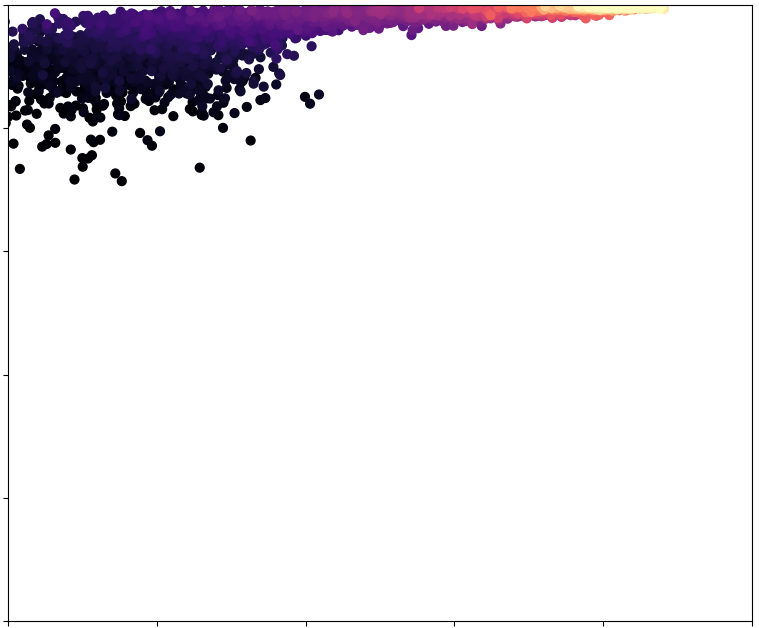}
        \label{fig:a}
        (c)
    \end{minipage}
    \begin{minipage}[b]{0.49\textwidth}
        \centering
        \includegraphics[width=\textwidth]{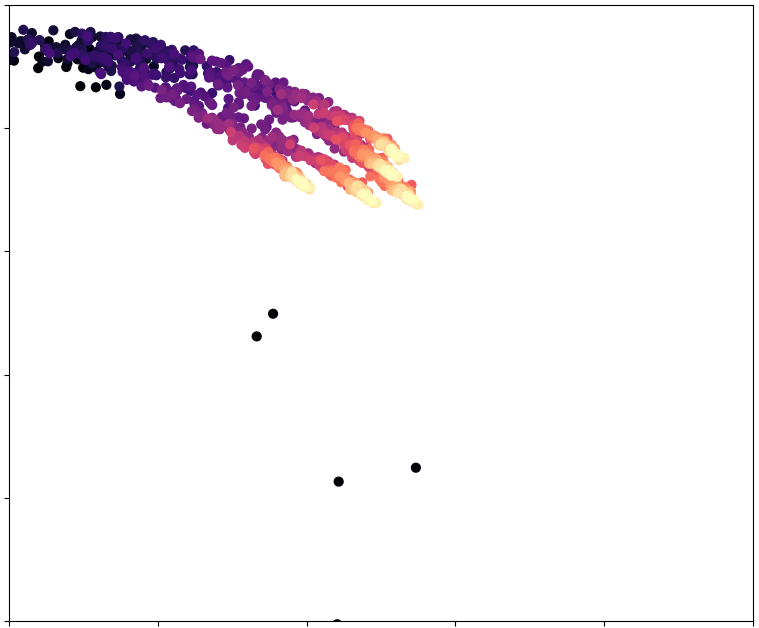}
        \label{fig:b}
        (d)
    \end{minipage}
    \begin{minipage}[b]{0.49\textwidth}
        \centering
        \includegraphics[width=\textwidth]{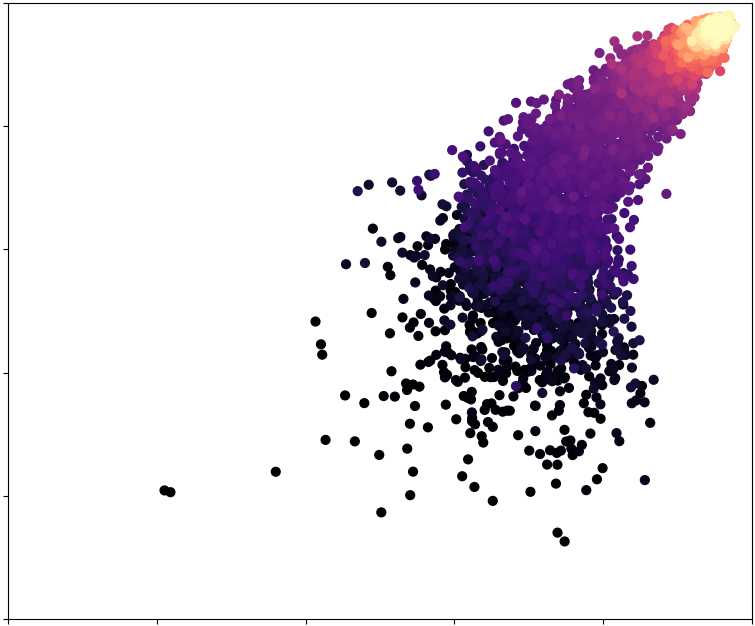}
        \label{fig:e}
        (e)
    \end{minipage}
    \begin{minipage}[b]{0.49\textwidth}
        \centering
        \includegraphics[width=\textwidth]{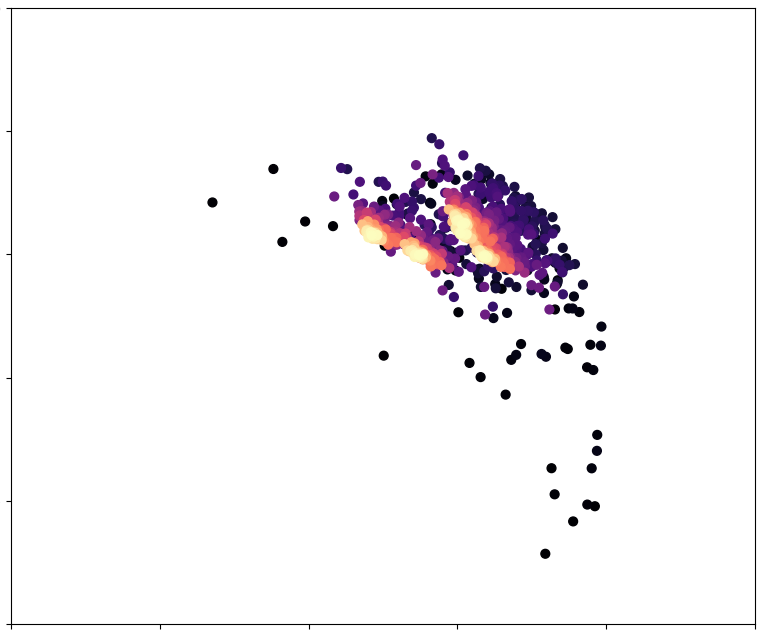}
        \label{fig:f}
        (f)
    \end{minipage}
    \caption{Color map of recall vs precision for different weight variations: The left column shows the progression of recall (x-axis) vs. precision (y-axis) during training, while the right column shows the corresponding values for the validation set. Lighter dots indicate later training epochs. The first row, \textbf{(a, b)}, corresponds to \textit{dynamic weights} (our approach),  \textbf{(c, d)} to \textit{weights estimated from the training set}, and \textbf{(e, f)} to \textit{equal weights} for both classes. The scale of precision and recall is [0.5, 1.0] in all plots. Figure best viewed in color.}
    \label{precision_recall_cmap}
\end{figure}

\section{Discussion}
The key novelties of our technique are two-fold. First, we use dynamic, as opposed to fixed, weights to allow our networks to learn a more robust balance between false positives and false negatives. Second, we separate different types of tasks involved in retinal vessel segmentation. The likelihood estimation step only focuses on capturing the vascular structure without worrying about the final segmentation. Conversely, the targeted prediction step relies on having a precomputed likelihood map with which to make the final predictions. We discuss specific properties of our approach below.

\subsection{Dynamic weights yield smoother likelihood maps}
In Fig.~\ref{fig:pmap_datasets} (columns b and c), we can see that using class weights computed from the training set almost binarizes the image. All the likelihoods are very close to either 0 (background) or 1 (vessel); thus, we have no any useful information with which to refine our estimate in the targeted prediction step. Any information about fainter pixels was lost. However, if we use dynamic weights, a higher percentage of pixels will be assigned non-trivial likelihood values (Fig.~\ref{fig:pmap_datasets}(d)), which will allow our second network to better classify ambiguous pixels. 




\subsection{Dynamic weights yields more balanced precision and recall}
As we can observe from the results (Tables~\ref{tab:drive} to \ref{tab:vevioFrames}), our method consistently achieves a good balance between precision and recall. This implies that our approach handles fainter or more ambiguous pixels, rather than ignoring them. In contrast, the state-of-the-art results for the DRIVE dataset \cite{laddernet:journals/corr/abs-1810-07810} report precision and recall values of $0.8593$ and $0.7896$, respectively. Our corresponding, more balanced values were $0.8284$ and $0.8163$. We can see even more of a difference on the VEVIO dataset (Tables~\ref{tab:vevioMosaics} and \ref{tab:vevioFrames}). For the mosaics, the precision and recall for the U-net with class weights were $0.5537$ and $0.6832$, respectively. However, our full, dynamic pipeline yielded values of $0.6573$ and $0.6739$. 



Our dynamic weights allow a network to settle on a local optimum that better balances precision and recall. As noted earlier, retinal images have a much higher percentage of background pixels (over 90\%). Thus, a network trained on this type of data will be more reticent to label an ambiguous pixel a vessel (since the prior probability is so much lower). This imbalance makes the network settle on a local optimum that is skewed towards high precision and low recall. Previous approaches have ameliorated this effect by assigning fixed class weights in order to re-balance the classification problem. For example, if only 10\% of pixels are vessels, then misclassifying those pixels will be penalized by a factor of 9. However, not all faint vessels are similar or distributed equally across different images, so a one-size-fits-all treatment is suboptimal. Our technique, on the other hand, applies \textit{stress tests} during the training process to ensure that the network is robust across different ratios of precision and recall. As our experiments show, this approach allows the network to settle on a a better optimum. 

In more detail, Fig.~\ref{precision_recall_cmap} illustrates the effects of dynamic vs. fixed weights during training for the DRIVE dataset. Here, each subplot on the left-hand side shows precision vs. recall during training, while the subplots on the right show this value for the validation set. As Fig.~\ref{precision_recall_cmap}(c) shows, a network that uses weights calculated from the training set will begin by strongly favoring recall, since a false negative is penalized more heavily than a false positive. In contrast, an unweighted network (Fig.~\ref{precision_recall_cmap}(e)) will favor precision at first, since the probability that an ambiguous pixel is a vessel is very low, \textit{a priori}. Over time, both networks learned to classify the training set, but their initial bias lead to poor results on the validation set, suggesting overfitting (Fig.~\ref{precision_recall_cmap}(d and f)). This behavior implies that the network was very confident about easy pixels and therefore settled on an optimum that favored this class of pixels; however, this choice made it ignore fainter pixels, in turn. Our method, on the other hand, (Fig.~\ref{precision_recall_cmap}(a, b)), had more balanced scores throughout the training process. The optimum on which it settled was more robust and thus achieved better scores on the validation and test sets. Our results suggest that the network was able to learn features that separate faint vessels from noise in the background.

\subsection{Effect of different weighing schemes}
Finally, we carried out additional analyses to investigate the impact of different weighing schemes on the final segmentation result. First, we trained a U-net architecture (without targeted prediction) with equal weights for the two classes. In other words, each misclassification, whether false positive or false negative, was weighted equally. Table~\ref{tab:equal_weights} shows the results of this equal weighing on our different datasets (we also show the class-weighted results for each dataset for comparison). Equal weights yielded better precision but worst recall. Due to the roughly 9 to 1 class imbalance between background and vessel pixels, a network trained with equal weights will tend to label ambiguous pixels as background. We verified this trade-off further by systematically varying the class weights for the DRIVE dataset and recording the resulting precision and recall values. Table~\ref{tab:different_weights_drive} lists the results when training with background-to-vessel weights of \{1,5\}, \{1,1\} (equal weights), \{5,1\}, and \{10,1\}. For comparison, the class-based weights were roughly 1 to 10. We also tested a U-net with dynamic weights, but with a range of weights spanning only 10 to 1 (our main experiments used a range of 100 to 1). In this case, the dynamic weights has less variance across training epochs.

Interestingly, the best fixed-weight result for the DRIVE dataset used equal weights (this result was not consistent across datasets, though, as Table~\ref{tab:equal_weights} shows). Overall, we can see that the trade-off of precision vs. recall is driven by the ratio of class weights. Also, the dynamic weights with a 10-to-1 range did not balance the precision vs. recall as much as the 100-to-1 weights, so their results more closely resembled equal weights. We observed a similar pattern for the other datasets (we omitted these results for conciseness). Skewed weights tend to favor either precision or recall and reduce the other score accordingly. By effectively balancing these two scores, our dynamic weights achieve the best result.


\section{Conclusion}
An accurate segmentation of the vascular structure is crucial for retinal disease diagnosis. However, this task is not easy because the quality and features of a retinal image depend on many factors, including the imaging device, lighting conditions, and an individual's anatomy. Thus, trying to enforce a single, static approach for all retinal images is suboptimal. Instead, by separating retinal vessel segmentation into sub-tasks, we have more degrees of freedom with which to adapt our processing to the current image. Furthermore, dynamic weights allow a network to learn to classify all types of vessels, regardless of their colour or intensity, which in turn generates a likelihood map with clear distinctions between vessel, background, and noise. Our targeted prediction step then uses this likelihood map to better classify ambiguous pixels. Our technique gives better results than state-of-the-art techniques, many of which are much complex and less intuitive. Instead of optimizing a single, black box, our approach breaks down the problem into more manageable steps and makes these steps more robust by using dynamic weights. In future work, we plan to apply our pipeline to other medical imaging domains, including CT and MRI scans. We also intend to investigate the theoretical properties of our stochastic weights further. 

 


\section*{Disclosures}
The authors declare that there are no conflicts of interest related to this article.


\bibliographystyle{osajnl}
\bibliography{bibliography}






\end{document}